\documentclass[10pt,times,mathptm,psfig,twocolumn,journals]{IEEEtran}
\usepackage{graphicx}
\usepackage{amsmath}
\usepackage{amssymb}
\usepackage{booktabs}
\usepackage{bm}
\usepackage{mathrsfs}
\usepackage{makecell}
\usepackage{cuted}
\usepackage{capt-of}
\usepackage{pifont}
\usepackage{color}
\usepackage{ulem}
\usepackage{cite}
\usepackage{hyperref}
\usepackage{colortbl}

\definecolor{Gray}{RGB}{210,210,210}
\definecolor{Black}{RGB}{0,0,0}
\definecolor{mygray}{gray}{.85}

\begin{document}

\title{Cylin-Painting: Seamless {360\textdegree} Panoramic Image Outpainting and Beyond}

\author{Kang~Liao$^{*}$, Xiangyu Xu$^{*,\dagger}$, Chunyu Lin$^{\dagger}$, Wenqi Ren, Yunchao Wei, Yao Zhao,~\IEEEmembership{Fellow,~IEEE}
\thanks{This work was supported by the National Natural Science Foundation of China (No.62172032, No.62120106009, No.62302385, No.U23A20314).}
\thanks{Kang Liao, Chunyu Lin, Yunchao Wei, and Yao Zhao are with the Institute of Information Science, Beijing Jiaotong University, China, and also with the Beijing Key Laboratory of Advanced Information Science and Network Technology, China (email: kang\_liao@bjtu.edu.cn, cylin@bjtu.edu.cn, wychao1987@gmail.com, yzhao@bjtu.edu.cn).}
\thanks{
Xiangyu Xu is with the School of Mathematics and Statistics, Xi'an Jiaotong University, China (email: xuxiangyu2014@gmail.com).}
\thanks{
Wenqi Ren is with the School of Cyber Science and Technology, Sun Yat-sen University, China (email: rwq.renwenqi@gmail.com).

\textit{$^{*}$Co-first author. $^{\dagger}$Corresponding author.}}
}

\maketitle
\graphicspath{{imgs/}}
\begin{abstract}
Image outpainting gains increasing attention since it can generate the complete scene from a partial view, providing a valuable solution to construct {360\textdegree} panoramic images. As image outpainting suffers from the intrinsic issue of unidirectional completion flow, previous methods convert the original problem into inpainting, which allows a bidirectional flow. However, we find that inpainting has its own limitations and is inferior to outpainting in certain situations. The question of how they may be combined for the best of both has as yet remained under-explored. In this paper, we provide a deep analysis of the differences between inpainting and outpainting, which essentially depends on how the source pixels contribute to the unknown regions under different spatial arrangements. Motivated by this analysis, we present a Cylin-Painting framework that involves meaningful collaborations between inpainting and outpainting and efficiently fuses the different arrangements, with a view to leveraging their complementary benefits on a seamless cylinder. Nevertheless, straightforwardly applying the cylinder-style convolution often generates visually unpleasing results as it discards important positional information. To address this issue, we further present a learnable positional embedding strategy to incorporate the missing component of positional encoding into the cylinder convolution, which significantly improves the panoramic results. It is noted that while developed for image outpainting, the proposed algorithm can be effectively extended to other panoramic vision tasks, such as object detection, depth estimation, and image super-resolution. 
Code will be made available at \url{https://github.com/KangLiao929/Cylin-Painting}.
\end{abstract}

\markboth{}
{Shell \MakeLowercase{\textit{et al.}}: Bare Demo of IEEEtran.cls for IEEE Transactions on Magnetics Journals}
\IEEEpeerreviewmaketitle

\section{Introduction}

\begin{figure}[!t]
  \centering
  \includegraphics[width=.46\textwidth]{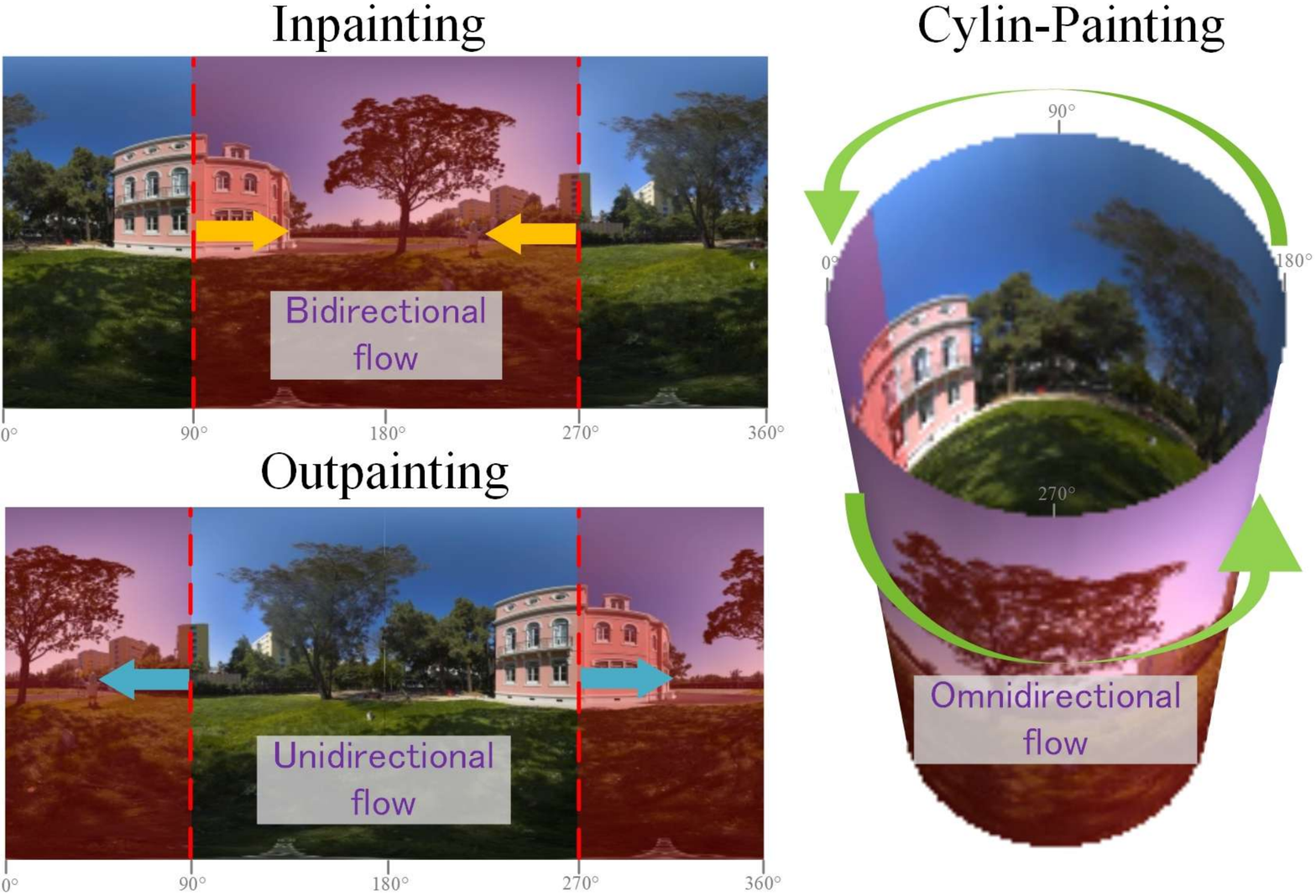} 
  \caption{Comparison of different spatial arrangements in inpainting, outpainting, and our Cylin-Painting. By leveraging the complementary benefits of inpainting and outpainting on a seamless cylinder, our method enables an omnidirectional flow to efficiently reconstruct semantically consistent scenes.} 
  \label{fig:teaser}
    %\vspace{-0.1cm}
\end{figure}

Image outpainting aims to build a semantically consistent extrapolation beyond the image boundaries~\cite{12,13,14,15, bowen2021oconet, khurana2021semie, li2021controllable, wang2021sketch, lu2021bridging}. In this work, we consider the problem of outpainting 360{\textdegree} panoramic images. This is an important problem that can be applied to virtual reality and numerous applications on wearable devices. The study of panoramic vision problems also has gained heightened interest in recent years, such as semantic segmentation~\cite{jaus2023panoramic, zhang2022bending, yang2021capturing, gao2022review}, object detection~\cite{de2018eliminating, chou2020360}, depth estimation~\cite{zhuang2023spdet, shen2022panoformer}, and optical flow estimation~\cite{shi2023panoflow}, etc. However, it is challenging to generate the panoramic context due to the large size of the missing regions and the unidirectionality of the information flow (Fig.~\ref{fig:teaser}). 

To address the above problem, Kim et al. \cite{kim2021painting} convert the outpainting problem into an inpainting one, which allows bidirectional information flow as shown in Fig.~\ref{fig:teaser}.
However, the inpainting method has its own issues and is inferior to outpainting in certain situations, while little literature has studied their relationships and limitations.

In this paper, we provide a deeper analysis that the difference between inpainting and outpainting essentially depends on how the source pixels contribute to the unknown locations under different \textit{spatial arrangements} of the input image. The inpainting excels at building the data relevance from two-sided information. 
In contrast, the outpainting better encourages a holistic understanding of the image structure, avoiding conflicts in the reconstruction process. 
Motivated by this, we present a Cylin-Painting framework that involves meaningful collaborations between the two families of image completion and efficiently fuses the different arrangements, to leverage their complementary benefits in a seamless cylinder. As shown in Fig.~\ref{fig:teaser}, our method offers an omnidirectional flow compared to the bidirectional flow in inpainting and unidirectional flow in outpainting. As a consequence, the proposed Cylin-Painting can reason the correlation between left and right boundaries, which enables semantically consistent 360{\textdegree} panoramic extrapolated results.

While the cylinder-style convolution can be efficiently implemented into panoramic images, directly exploiting it often leads to inferior results with disordered structures. 
Such degeneration is somewhat surprising considering the improved information flow of Cylin-Painting. 
To investigate this issue, we perform a comprehensive analysis of the feature maps and find that the problem is mainly caused by the missing positional information.
Compared to the vanilla convolution, the cylinder-style convolution unintentionally discards the vertical positional encoding of CNN feature maps. As a consequence, the neural network fails to build the spatial correlation between the original content and the blank region, leading to visually unpleasant results.
Based on this analysis, we present a new learnable positional embedding strategy to incorporate the missing component of positional encoding into the cylinder convolution, significantly improving the extrapolation results on the 360{\textdegree} panoramic image.

Extensive experiments on three public panoramic image datasets demonstrate the effectiveness of our approach. While developed based on image outpainting, the proposed solution can be effectively and flexibly applied to other panoramic vision problems, such as object detection, depth estimation, and super-resolution. 

Our main contributions can be summarized as follows.

\begin{itemize}
    \item To our knowledge, we are the first effort to analyze the essential difference between image inpainting and image outpainting theoretically and experimentally.
    \item By analyzing the essential difference between inpainting and outpainting, we propose a Cylin-Painting to efficiently fuse the different spatial arrangements of the input image, which also enables high-quality seamless 360{\textdegree} panoramic image extrapolation. 
    \item We make an early attempt to systematically describe the strengths and limitations of positional encoding in CNNs. Furthermore, we tame the cylinder convolution with a novel learnable positional encoding, which essentially improves the generation results.
    \item Our method can serve as a plug-and-play module and flexibly extend to other 360{\textdegree} panoramic vision tasks, including both low-level and high-level ones.
\end{itemize}

\section{Related Work}\label{sec2}

\subsection{Image Completion} 
Image completion is to generate the missing parts in incomplete images, it mainly includes the inpainting \cite{15,16,17,18,19,20,zeng2020high} and the outpainting \cite{12,13,14,15,bowen2021oconet,khurana2021semie,li2021controllable,wang2021sketch,lu2021bridging}. In general, image outpainting is more challenging than inpainting because of the unidirectionality of the information flow. Sabini et al. \cite{111} first proposed the learning-based outpainting with the powerful generation ability of the generative adversarial networks (GANs). SRN \cite{13} was presented to gradually improve the outpainting performance using a feature expansion module and content prediction module. Teterwak et al. \cite{12} conditioned the discriminator by pre-trained features from the InceptionV3 network, which helps the extrapolated results to match the original scene in the semantics space. A spiral generative network \cite{29} is designed to conduct the image extrapolation following the perception fashion of humans. In addition to the normal field-of-view (FoV) image, FisheyeEX~\cite{liao2022fisheyeex} and FlowLens~\cite{shi2022flowlens} explore to further extend the scene's FoV, designing polar coordinate-based outpainting strategy and flow-guided clip-recurrent transformer for fisheye images, respectively. FishDreamer~\cite{shi2023fishdreamer} is then proposed to produce semantically consistent visual content beyond the fisheye FoV. This is achieved by collaboratively learning both the scene's semantics and pixel patterns, while taking into account the fisheye's polar distributions. There are some recent works designed for panoramic image outpainting~\cite{somanath2021hdr, hara2021spherical, akimoto2022diverse, hara2022spherical}. SIG-SS~\cite{hara2021spherical, hara2022spherical} generates a spherical image from a few normal-field-of-view (NFoV) images while controlling the symmetry of generated scenes. EnvMapNet~\cite{somanath2021hdr} trains a neural network by adjusting the pixel loss weights to consider variations in latitudinal information density resulting from the projection. OmniDreamer~\cite{akimoto2022diverse} leverages a transformer for diverse scene modeling and improves the properties of generated panoramic images. Although some works considered panoramic consistency, the significance of preserving positional information has remained unexplored.

\subsection{Positional Encoding} 
With the success of Transformer-based networks over the past few years, more computer vision tasks have gained improvements over CNNs. Positional encoding is a crucial aspect of transformers, given their lack of inherent positional awareness. In the context of panoramic images, where spatial relationships between features are vital, some panoramic vision works~\cite{jiang2022lgt, li2023panoramic, zheng2023both} introduce panoramic-oriented positional encodings to compensate for the panoramic characteristics. Compared with these transformer-based solutions, our method revisits the positional perception of \textit{CNN architectures}. While the cylinder-style convolution considers the left-right consistency of the panoramic image, we found it unintentionally discards the positional information and results in disordered generated images. Recent efforts also demonstrated that zero padding in CNNs can leak the positional information, enabling an implicit location-aware bias in the distribution of feature maps~\cite{islam2020much,alsallakh2020mind, kayhan2020translation,xu2021positional}. Applying positional encoding in transformer-based solutions is indispensable and intuitive. Instead, our proposed learnable positional embedding is first motivated by the failure cases from cylinder-style convolution, and is then designed based on a comprehensive analysis of how CNNs learn the positional information. It collaborates with cylinder-style convolution to boost realistic and visually pleasant outpainting results.

\begin{figure}[!t]
  \centering
  \includegraphics[width=.48\textwidth]{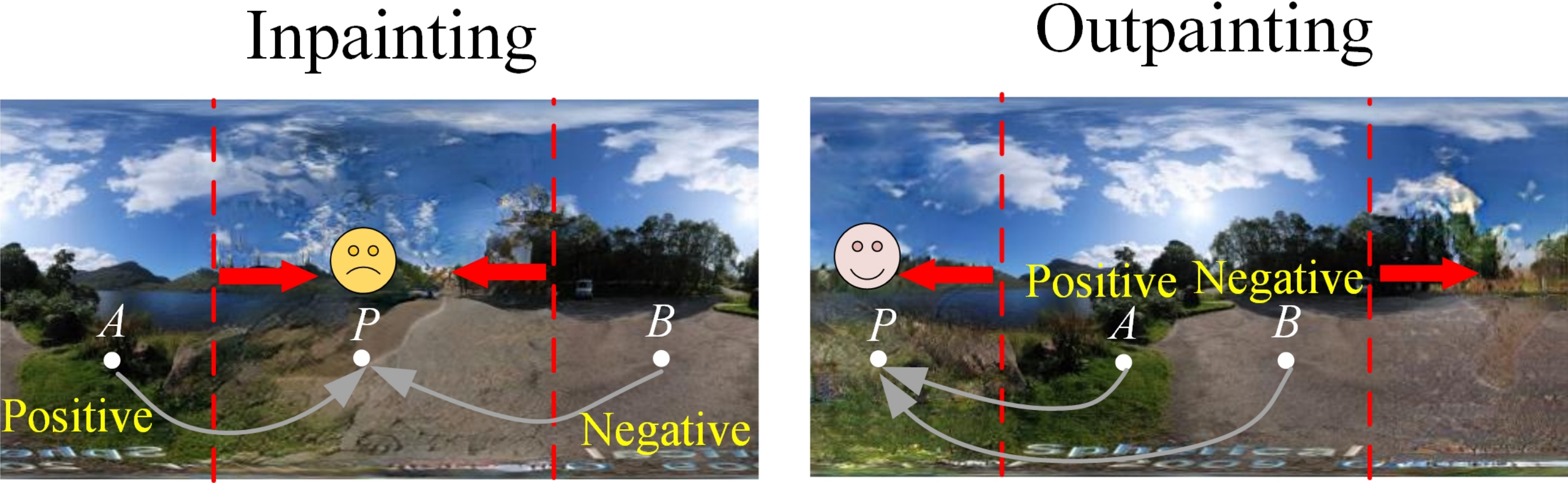} 
  \caption{A toy example where outpainting outperforms inpainting. The filling pixel $P$ is supposed to belong to the grass. The positive point $A$ has a high relevance to $P$ while the negative point $B$ with irrelevant semantics can confuse the scene reconstruction.} 
  \label{fig:toy-example}
    %\vspace{-0.2cm}
\end{figure}

\section{Preliminaries}\label{sec3}
\label{s3}

\subsection{Problem Statement}
\label{s3.1}
Given a narrow field-of-view image $I_N \in \mathbb{R}^{\theta \times \varphi \times 3}$, our objective is to extrapolate semantically-consistent contents from both left and right boundaries, seamlessly reconstructing a 360{\textdegree} panoramic image $I_P \in \mathbb{R}^{\theta^{\prime} \times \varphi \times 3}$ in an equirectangular projection (ERP) format \cite{su2017learning}. In particular, $\theta$ is the azimuthal angle, and $\varphi$ is the polar angle, where $0 < \theta < \theta^{\prime} = 2\pi$.

Previous methods study inpainting and outpainting in isolation. Many researchers believe outpainting is more challenging than inpainting due to its one-side constraint, i.e., the information flow from source pixels to unknown target pixels is unidirectional. However, little literature comprehensively studied this point with theoretical or experimental analysis. \textit{Dose the inpainting outperform the outpainting in all situations?} 
Interestingly, we find the answer to this question is non-affirmative.
In Fig.~\ref{fig:toy-example}, we show an example in which the outpainting outperforms the inpainting. The `positive' in Fig.~\ref{fig:toy-example} indicates a source pixel that has a close relationship to the target pixel, which helps the content reconstruction. The `negative' indicates a source pixel irrelevant to the target pixel. To be more specific, for a filling pixel $P$ (supposed to be grass), the pixels $A$ and $B$ have similar distances and provide bidirectional information in inpainting. However, this spatial arrangement cannot represent which pixel is more important to reconstruct the pixel $P$ due to the close distances. By contrast, the irrelevance of pixel $B$ to pixel $P$ can be well controlled in unidirectional outpainting as the distance increases. Motivated by this observation, we further analyze the essential relationship between inpainting and outpainting as below.

\begin{figure}[!t]
  \centering
  \includegraphics[width=.5\textwidth]{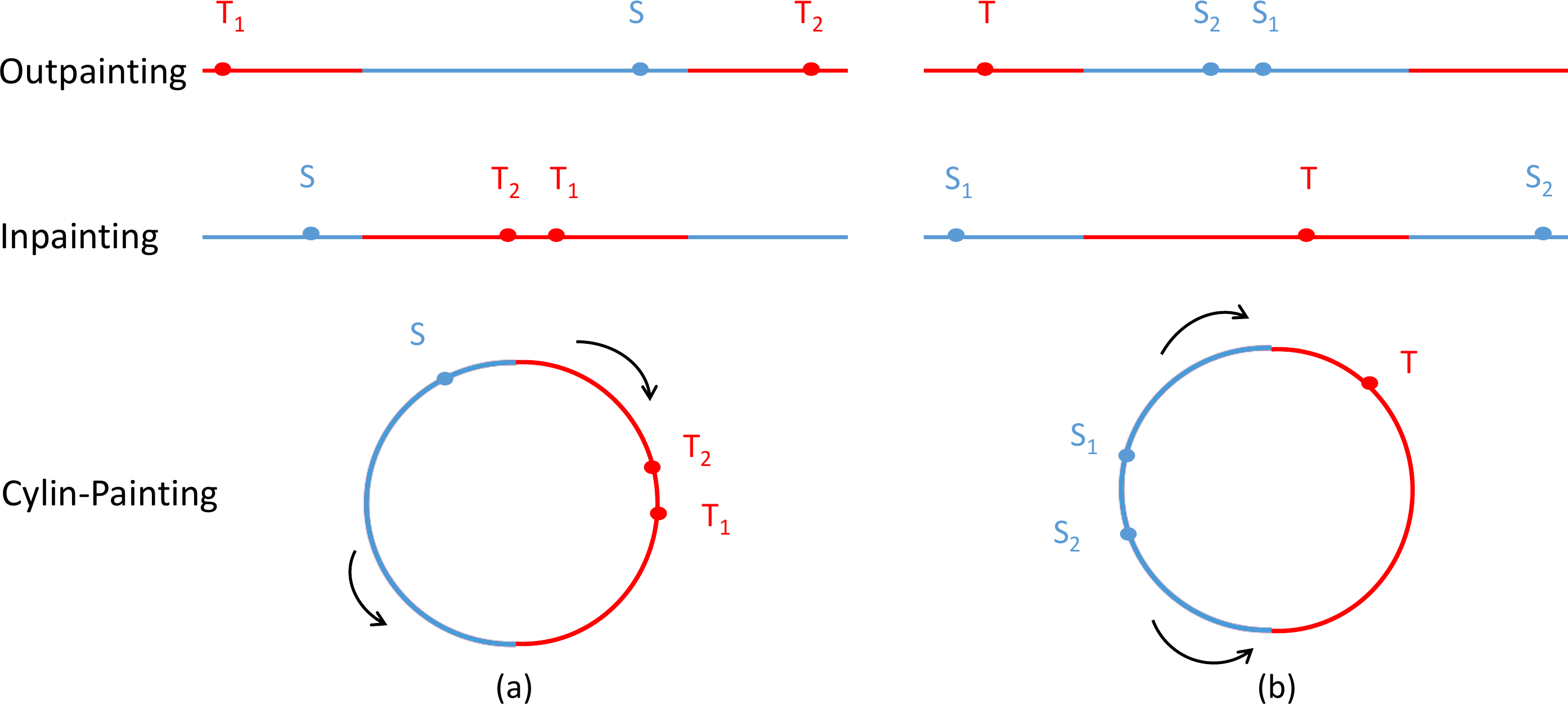} 
  \caption{Comparing spatial arrangements of outpainting, inpainting, and our Cylin-Painting. The blue and red segments represent the known and unknown regions in reconstruction, respectively.} 
  \label{fig:spatial-arrangement}
    %\vspace{-0.2cm}
\end{figure}

\subsection{Spatial Arrangement}
\label{s3.2}
In this part, we give a deep analysis of inpainting and outpainting, and illustrate that the difference between them essentially depends on how the source pixels contribute to the unknown locations under different \textit{spatial arrangements} of the input image. By modeling the different spatial arrangements in Fig.~\ref{fig:spatial-arrangement}, we show our Cylin-Painting can combine the advantages of both inpainting and outpainting.

Suppose that $S$ and $T$ indicate a source pixel and a target pixel, respectively. Due to the translation equivariance of CNNs, the influence of $S$ to $T$ can be written as a function of their relative position:
\begin{equation}%\label{eq1}
r_{S \rightarrow T} = \boldsymbol{f}(\theta_T - \theta_S),
\end{equation}
where $\theta_T$ and $\theta_S$ are the horizontal positions of pixel $T$ and $S$, which indicate their azimuthal angles in ERP.

As shown in Fig.~\ref{fig:spatial-arrangement}(a), while $T_1$ and $T_2$ are neighboring pixels, the outpainting leads to an abrupt change between $r_{S \rightarrow T_1}$ and $r_{S \rightarrow T_2}$, resulting in discontinuous completion results.
While converting the spatial arrangement to inpainting solves this problem, this solution has its own issue as shown in Fig.~\ref{fig:spatial-arrangement}(b): while $S_1$ and $S_2$ are neighboring pixels, the inpainting leads to an abrupt change between $r_{S_1 \rightarrow T}$ and $r_{S_2 \rightarrow T}$, which confuses the reconstruction process and hinders the neural network from grasping the whole content distribution of the input effectively.

Instead, the proposed Cylin-Painting can address the above two limitations by:
\begin{equation}\label{eq2}
r_{S \rightarrow T} = \boldsymbol{f}(\theta_T - \theta_S) + \boldsymbol{f}(2\pi - (\theta_T - \theta_S)),
\end{equation}
which well guarantees the following properties:
\begin{itemize}
    \item A source pixel has a continuous influence on neighboring target pixels. 
    \item A target pixel is continuously influenced by neighboring source pixels.
\end{itemize}

\section{Cylin-Painting}\label{sec4}
Previous methods only convert outpainting into inpainting or vice versa, without leveraging their complementary benefits.
A straightforward way to combine these two techniques is using a two-branch network to separately conduct inpainting and outpainting and then fusing the outputs from both branches, which, however, is not as effective and concise as the proposed Cylin-Painting as will be shown next.
In this section, we explore how to realize the concept of Cylin-Painting to bridge the gap between inpainting and outpainting, and present a compact layer to efficiently fuse the different spatial arrangements.

\subsection{Cylinder-Style Convolution}
\label{s4.1}
As described above, inpainting and outpainting have different effects on reconstructing a missing pixel, and their different spatial arrangements are complementary to each other. Therefore, we propose a cylinder-style convolution layer to realize meaningful collaborations between these two kinds of image completion methods.

Specifically, suppose that the input feature map $\mathcal{F}_{in} \in \mathbb{R}^{\Theta \times \Phi \times C}$ is processed by $C^{\prime}$ convolutional kernels: $\{\mathcal{K}^t \in \mathbb{R}^{(2M + 1) \times (2N + 1) \times C}, t=1,\cdots,C^{\prime} \}$. 
The cylinder-style convolution layer can be written as:
\begin{equation}\label{eq3}
\begin{split}
\mathcal{F}_{out}(\theta, \varphi, t) =& (\mathcal{K}^t * \mathcal{F}_{in})(\theta, \varphi) \\
=& \sum_{c=1}^{C}\sum_{m=-M}^{M}\sum_{n=-N}^{N} \mathcal{K}^t(m + M, n + N, c) \cdot \\ &\mathcal{F}_{in}(\mathcal{M}{(s_\theta  \theta - m)}, s_\varphi  \varphi - n, c),\\
\end{split}
\end{equation}
where $\mathcal{F}^t_{out}$ is the $t$-th feature map of the output $\mathcal{F}_{out} \in \mathbb{R}^{\Theta^{\prime} \times \Phi^{\prime} \times C^{\prime}}$. $*$ is the convolution operation. $s = [s_\theta, s_\varphi]$ represents the stride. $\mathcal{M}$ denotes the modulo division and can be efficiently realized with circular padding.

Previous methods use the vanilla convolution that pads the boundary with zero values, leading to a seamed outpainting effect on the panoramic image. 
Instead, by building connections between the left and right contents on the circular surface, our Cylin-Painting enables a seamless and consistent extrapolation result.

\begin{figure*}[t]
\centering
\includegraphics[width=1\linewidth]{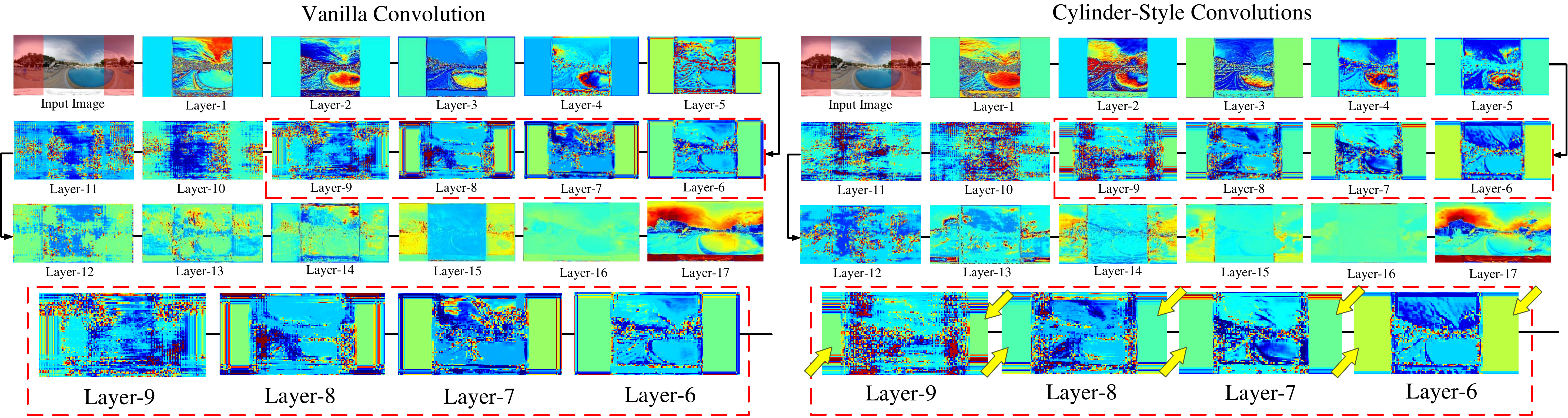}
\caption{Visualization of the feature maps in a vanilla convolutional network (left) and cylinder-style convolutional network (right). The red dotted rectangle shows a zoom-in of the feature maps from Layer-6 to Layer-9. The yellow arrow highlights the missing vertical lines in Cylin-Painting compared to the vanilla convolution.}
\label{fig:feature_maps}
%\vspace{-0.3cm}
\end{figure*}

\subsection{Network Architecture}
\label{s4.2}
Similar to existing conditional image generation methods, the proposed Cylin-Painting is based on a GAN framework which consists of a generator network $G$ and a discriminator network $D$. Specifically, the generator shares a similar architecture with the U-Net, which takes an incomplete image and a mask as the input and produces the reconstructed panoramic image. The cylinder-style convolution is used to replace each convolutional layer in the U-Net, where the gating mechanism~\cite{yu2019free} is also incorporated. In addition, we use the spectral normalization \cite{miyato2018spectral} to normalize all layers in $D$ to satisfy the Lipschitz constraints of Wasserstein GANs \cite{arjovsky2017wasserstein}.

\section{Positional Encoding in Cylin-Painting}\label{sec5}

As described above, we exploit a cylinder-style convolution to combine the benefits of inpainting and outpainting. Although such a design enables a seamless and consistent completion result across image boundaries, we empirically find that it may lead to inferior visual appearances with disordered image structures.
In this section, we unveil the cause of this issue and propose a novel method to address it.

\subsection{Feature Map Visualization}\label{s5.1}

An intuitive way to investigate how neural networks work is feature map visualization. 
By comparing the feature map of each layer of the vanilla convolutional network and cylinder-style network in Fig.~\ref{fig:feature_maps}, we can observe several intriguing phenomenons that help explain why the performance degrades for a straightforward implementation of Cylin-Painting: 
(i) The outpainting process is progressive: the empty region is filled layer by layer till the end of the encoder network (Layer-10), and the decoder part starting from Layer-11 mainly improves the details of the completed image. (ii) In the feature maps, some straight-line patterns emerge and gradually expand from the edge to the center region. (iii) Compared to the outpainting network with the vanilla convolution, the cylinder-style convolution does not have vertical lines in feature maps. 
Therefore, we hypothesize that the potential reason behind the decreased performance is the missing vertical line patterns.

\subsection{Line Patterns}\label{s5.2}

Based on the above observations, we raise the following two questions: where the lines come from, and what the lines represent, which are separately answered below.

\vspace{1mm}
\noindent \textbf{The line patterns are accidentally introduced by zero padding.}
Zero padding is an essential component of modern CNN networks, \textit{e.g.} in VGG~\cite{simonyan2014very} and ResNet~\cite{he2016deep}.
It extends the feature map with zero values and is mainly used for maintaining the size of the feature maps throughout the network.
The padding size is often chosen based on the kernel size, stride, and dilation rate. 
With zero padding, when a convolutional kernel extracts the feature at the boundary of an image, the abrupt changes between the zero values and original RGB information are mistakenly recognized as an edge. 
For brevity, we consider a 1D convolution between the left-side vertical lines $\mathcal{L}_{n} \in \mathbb{R}^{\Theta \times p}$ and $C^{\prime}$ 1D convolutional kernels $\{\mathcal{K}^t \in \mathbb{R}^{(2M + 1) \times  C}, t=1,\cdots, C^{\prime}\}$.
By applying a left-side zero padding $\mathcal{Z} \in \mathbb{R}^{\Theta \times M}$, we can obtain the line patterns $\mathcal{L}_{n+1} \in \mathbb{R}^{\Theta \times (p+M)}$ in the next feature map:

\begin{equation}\label{eq1}
\begin{split}
\mathcal{L}_{n+1}(\theta, \varphi,t) =& (\mathcal{K}^{t} * (\mathcal{Z} \mathbin\Vert \mathcal{L}_{n}))(\theta, \varphi+M)\\
=& \sum_{c=1}^{C}\sum_{m=-M}^{M} \mathcal{K}^t(m + M, c) \cdot \\ &(\mathcal{Z} \mathbin\Vert \mathcal{L}_{n})(\theta , \varphi + M - m, c),\\
\end{split}
\end{equation}
where $\mathbin\Vert$ indicates the concatenation operation at the width dimension. The horizontal lines can also be described using the above equation by changing the convolution and concatenation from 1D to 2D.
As the number of convolutional layers increases, the edge effect or line patterns will be increasingly more noticeable with more zero padding involved.

In contrast, the proposed cylinder-style convolution naturally avoids zero padding in the horizontal direction by introducing the modulo operation $\mathcal{M}$ in Eq.~\ref{eq3}.
Therefore, the feature maps of Cylin-Painting do not have the vertical line patterns in Fig.~\ref{fig:feature_maps}.

\vspace{1mm}
\noindent \textbf{The line patterns encode positional information.}
From Fig.~\ref{fig:feature_maps}, we can find that the pixel values in the line patterns are decided by spatial locations.
Hence, these patterns can be seen as a kind of positional marks indicating relative positions in shallow network layers. Moreover, the horizontal and vertical lines can interweave into a dense grid $\mathcal{G}$ at the deep convolutional layers such as Layer-10 and Layer-11, which potentially provides an implicit positional encoding for global spatial perception:

\begin{equation}%\label{eq2}
\mathcal{G}(\theta, \varphi) = \mathcal{L}^h(\theta, \varphi) \oplus \mathcal{L}^v(\theta, \varphi),
\end{equation}
where $\oplus$ denotes the element-wise addition of the horizontal and vertical lines. 
In other words, except for maintaining the size of the feature map, zero padding unintentionally introduces the position information into the feature map, which is beneficial for image reconstruction.
This explains our empirical observation that a straightforward version of Cylin-Painting without zero padding leads to degraded results.
We refer to the position encoding induced by the zero padding in CNNs as the padding position encoding (PPE).

\subsection{Zero Padding}\label{s5.3}

Except for positional encoding, our extensive experiments show that there are more underlying strengths and limitations behind the zero padding.

\begin{figure}[t]
\centering
\includegraphics[width=1\linewidth]{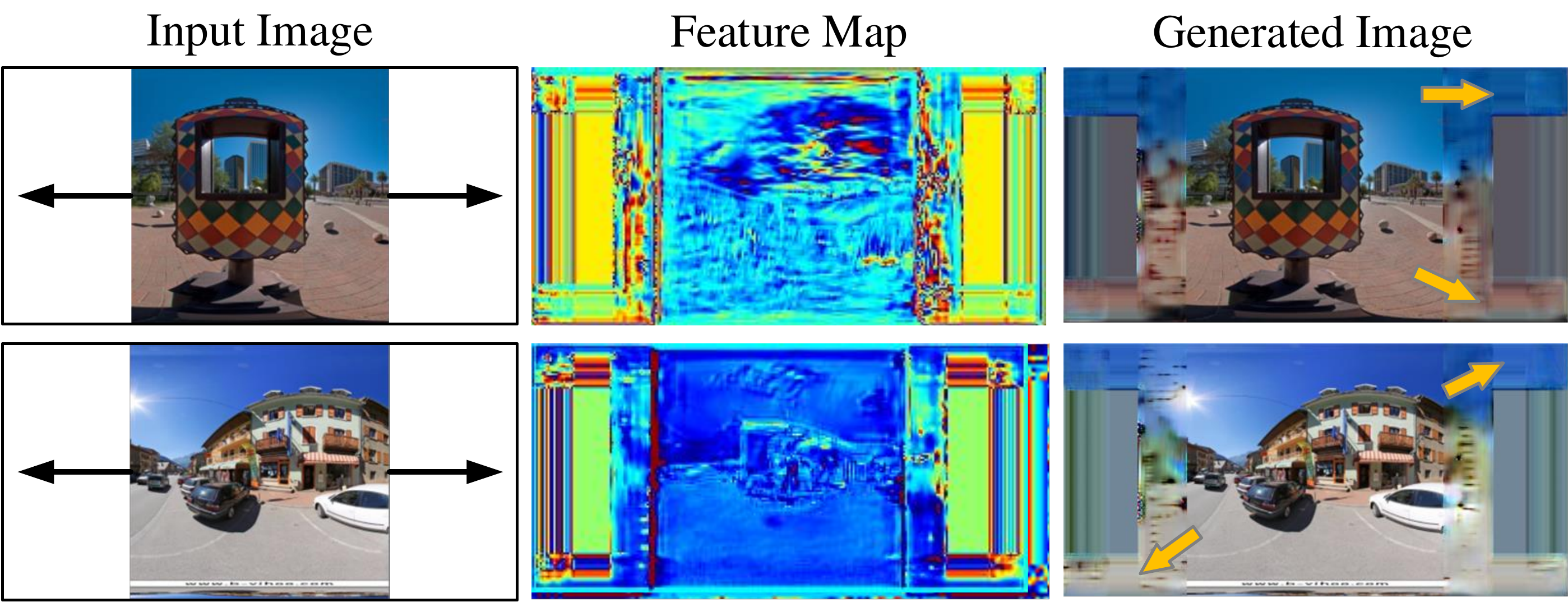}
\caption{Pigment for a conditional generation. Neural networks can generate plausible content from line patterns. Yellow arrows highlight the consistent semantics.}
\label{fig:pigments}
%\vspace{-0.3cm}
\end{figure}

Random noise is usually used for image generation in GANs. As a typical example, for StyleGAN \cite{karras2019style}, random latent vectors are mapped into different styles, which is shown to have the capability of controlling the generation results. Similarly, we find that the zero padding can also provide the latent vectors or \textit{pigments} for our conditional generation task. 
The neural network learns to rearrange these pigments and construct a plausible distribution, thus relieving the burden of generating the contents from zero. To experimentally verify this finding, we set the dilation rate of all convolutional layers to 1, which leads to some empty regions without line patterns in the last feature maps of the decoder. As shown in Fig.~\ref{fig:pigments}, the generative network can recover consistent semantics from these line patterns. By contrast, the unfilled regions without the line patterns still remain invalid in the generated image.

\begin{figure}[t]
\centering
\includegraphics[width=.95\linewidth]{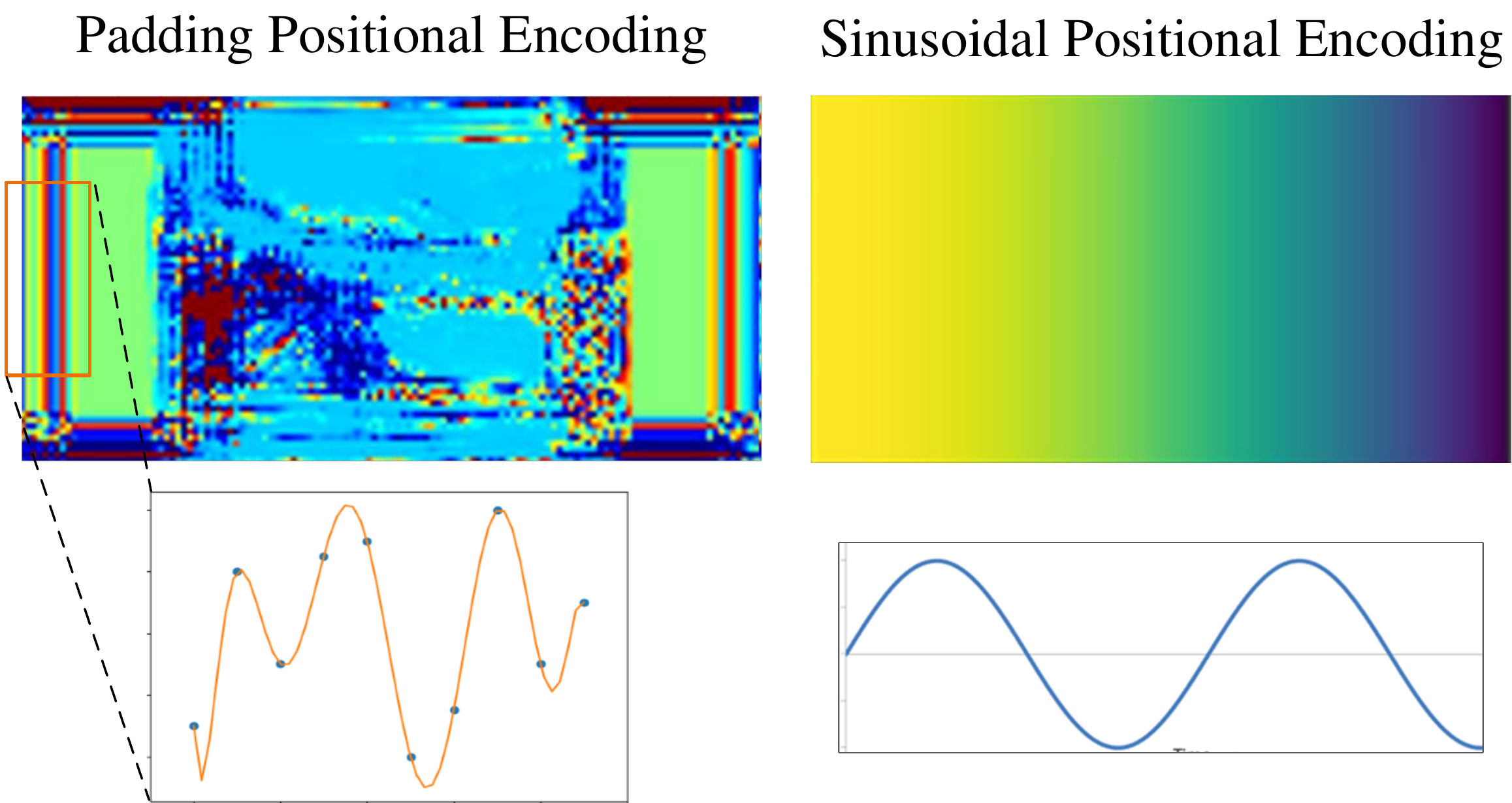}
\caption{Comparison of the padding positional encoding (PPE) and sinusoidal positional encodings (SPE).}
\label{fig:PPE}
%\vspace{-0.3cm}
\end{figure}

While the zero padding shows potential benefits for positional encoding and conditional generation, we argue its limitations as follows. First, compared to the sinusoidal positional encodings (SPE) in Transformers, the position information leaked by zero padding (PPE) is implicit and spatially unbalanced as shown in Fig.~\ref{fig:PPE}. In particular, we visualize different responses in the line patterns by fitting a polynomial model. As we can notice, SPE describes a more explicit and more stable spatial variance than PPE. Moreover, due to the fixed receptive field of the convolutional kernel, PPE grows slowly at the shallow layers in a neural network. Consequently, the unbalanced distribution of PPE cannot facilitate sufficient positional perception in CNNs. Third, our experiments demonstrate that zero padding tends to destroy local details in feature maps, which decreases the performances on low-level vision tasks such as image super-resolution. In the view of signal processing, the line artifact induced by zero padding can be regarded as additive high-frequency noise. It weakens the initial response on feature maps and violates the principle of detail preservation in low-level vision tasks. More experimental results will be shown in Section \ref{s6.4}.

\begin{figure*}[t]
\centering
\includegraphics[width=1\linewidth]{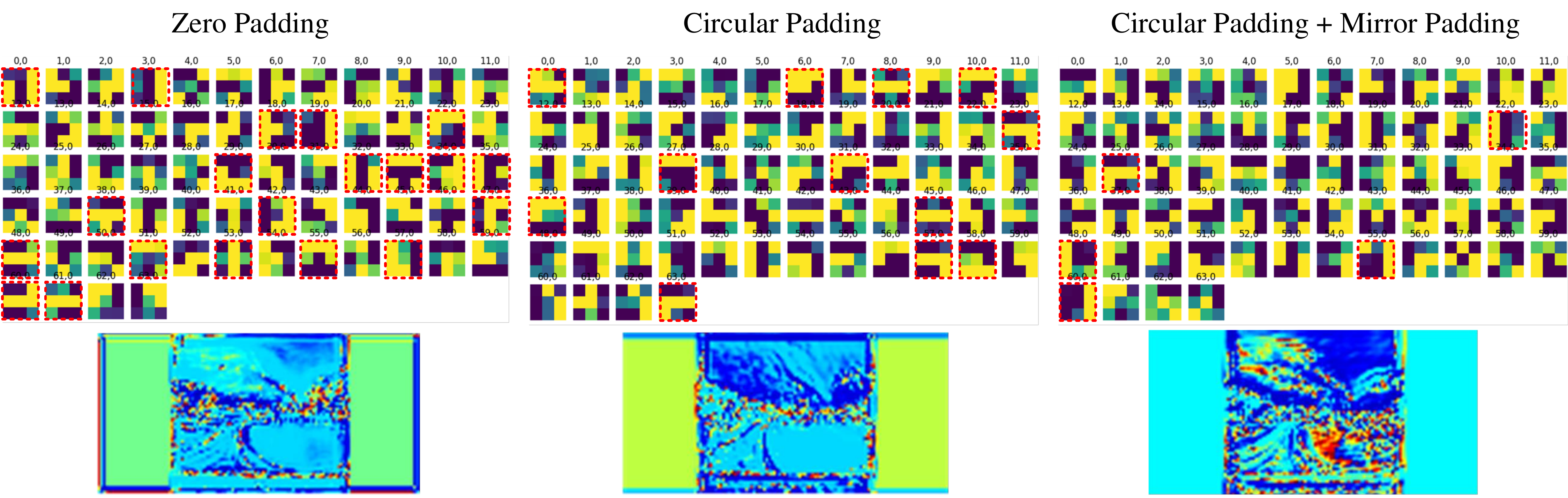}
\caption{Influence of padding strategy on the convolutional kernel. Red dotted boxes highlight some kernels with high weights in horizontal or vertical structures. Zero padding introduces the most line artifacts on the feature map, and thus the patterns of learned kernels are less diverse than others.}
\label{fig:kernel}
%\vspace{-0.3cm}
\end{figure*}

Finally, zero padding may influence the patterns of learned convolution kernels. In the implementation, we exploit three padding strategies for the same network: zero padding (both vertical and horizontal), circular padding (vertical), circular padding (vertical) + mirror padding (horizontal). 
We visualize the convolutional kernels and the feature maps in Fig.~\ref{fig:kernel}. It is evident that more horizontal and vertical patterns exist in kernels with the zero padding setting. The reason is that zero padding introduces more line patterns into the feature map, which are considered as one part of training data and further influence the kernel weights. We also compute the specific numbers of different kernel patterns as listed in Table \ref{tab:kernel}. Note that the numbers of introduced line patterns are descending from zero padding, circular padding, to circular padding + mirror padding. We have two observations: (i) In contrast to zero padding, circular padding enables fewer kernels with vertical patterns due to the discarded vertical lines. (ii) The kernel's pattern is the most varied in the C + M case, where the feature maps contain no line artifacts. 

\begin{table}\small
  \caption{Different kernel patterns induced by different padding strategies: Zero padding, circular padding, and circular padding + mirror padding (C + M). In these strategies, the numbers of introduced line artifacts are descending. More horizontal and vertical patterns indicate fewer varieties (Others) of the convolution kernels. Zero padding can introduce the most line artifacts into the feature map, which constrains the diversity of kernels.}
  \label{tab:kernel}
	 \centering
   \begin{tabular}{c|ccc}
    \hline
    Types & Zero & Circular & C + M \\
    \hline
    \hline
    Horizontal (H)  & 11 & 15& 2\\
    \hline
    Vertical (V) & 12 & 5 & 8\\
    \hline
    H + V & 23 & 20 & 10 \\
    \hline
    Others & 41 & 44 & \textbf{54} \\
  \hline
\end{tabular}
%\vspace{-0.3cm}
\end{table}

To summarize, we highlight several properties of the zero padding:

\begin{itemize}
    \item[$\bullet$] Strengths: 
    \begin{itemize}
    \item Maintaining the size of the feature map.
    \item Providing positional encoding.
    \item Pigment for a conditional generation.
    \end{itemize}
    \item[$\bullet$] Limitations:
    \begin{itemize}
    \item Implicit positional representation.
    \item Unbalanced spatial distribution.
    \item Additive noises for local details.
    \item Restricting diversity of the kernel patterns.
    \end{itemize} 
\end{itemize}

Based on the aforementioned discussions, we investigate a new positional embedding method below, which addresses the limitations while retaining the advantages of zero padding.

\begin{figure}[!t]
  \centering
  \includegraphics[width=.48\textwidth]{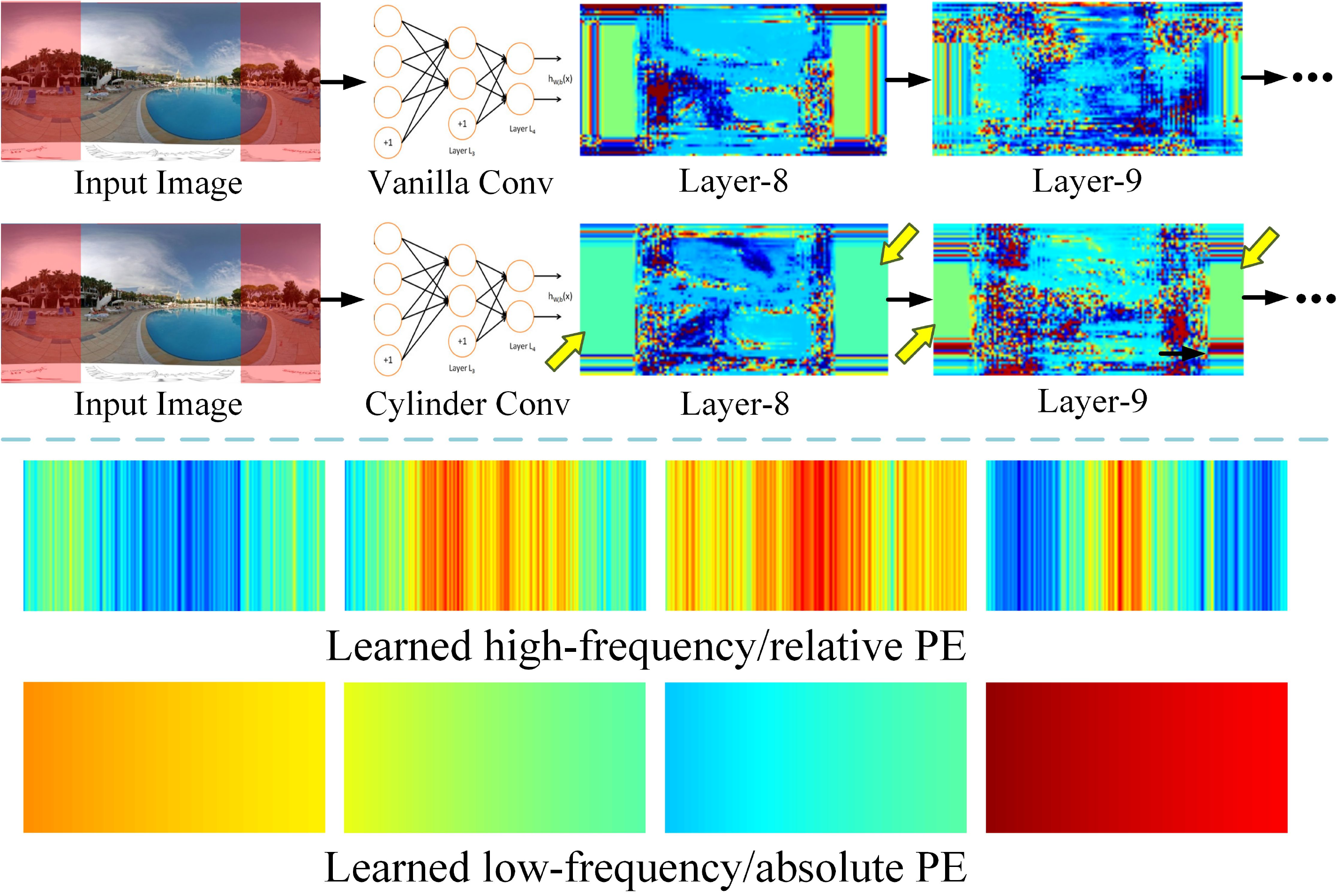} 
  \caption{Visualizations of the positional encoding (PE) in CNNs (top) and our learned positional encodings (bottom). Zero padding used in vanilla convolution introduces horizontal and vertical line artifacts into feature maps, which offer a relative position prior to the neural network. While the cylinder convolution enables complementary spatial arrangements, it discards the horizontal/azimuthal positional encoding as marked by yellow arrows. By contrast, our learnable positional embedding can obtain an explicit, compact, and evenly-distributed PE.} 
  \label{fig:SPE}
  %\vspace{-0.2cm}
\end{figure}

\subsection{Learnable Positional Embedding}
\label{s5.4}
Although PPE is automatically learned by CNNs, it cannot represent clear and compact position information. In addition, PPE could influence the local details on the feature map and the pattern variety of convolutional kernels. Inspired by the hand-crafted sinusoidal positional encoding (SPE) in the transformer, we present a learnable positional embedding that can be effectively integrated into the cylinder-style convolution.

\begin{table*}%\small
\begin{center}
\caption{Quantitative evaluation of the 360{\textdegree} panoramic image extrapolation results obtained by different methods, in which all methods are evaluated on three panoramic datasets: SUN360 \cite{xiao2012recognizing} (Left), Matterport3D \cite{chang2017matterport3d} (Middle), and 360SP \cite{chang2018generating} (Right). Note that FID (marked with underline) is generally regarded as a crucial metric for the image generation task.}
\label{table:1}

\begin{tabular}{p{2.11cm}<{\centering}|p{1.15cm}<{\centering}p{1.15cm}<{\centering}p{1.15cm}<{\centering}|p{1.15cm}<{\centering}p{1.15cm}<{\centering}p{1.15cm}<{\centering}|p{1.15cm}<{\centering}p{1.15cm}<{\centering}p{1.15cm}<{\centering}}

\hline
\    \ & \uline{FID} $\downarrow$  & PSNR $\uparrow$ & SSIM $\uparrow$ & \uline{FID} $\downarrow$  & PSNR $\uparrow$ & SSIM $\uparrow$ & \uline{FID} $\downarrow$  & PSNR $\uparrow$ & SSIM $\uparrow$\\
\hline
\hline

SRN \cite{13} & 59.40 & 17.79 & \textbf{0.70} & 42.56 & 16.62 & 0.72 & 46.48 & 17.41 & 0.69\\
\hline
NS-OP \cite{14} & 68.27 & 16.12 & 0.61& 79.82 & 16.81 & 0.67 & 79.27 & 17.11 & 0.63\\
\hline
Boundless \cite{12} & 53.43 & 17.39 & 0.64 & 37.13 & 17.32 & 0.62 & 30.12 & 17.22 & 0.64\\
\hline
SpiralNet \cite{29} & 55.52 & 17.30 & 0.64 & 34.10 & 17.01 & 0.69 & 33.97 & 17.61 & 0.68\\
\hline
Out2In \cite{kim2021painting} & 51.29 & 17.57 & 0.65 & 33.13 & 17.21 & 0.71 & 37.48 & 17.96 & 0.69\\
\hline
\cellcolor{mygray} \textbf{Ours} &
\cellcolor{mygray}{\textbf{42.13}} & \cellcolor{mygray}{\textbf{17.86}} & \cellcolor{mygray}{0.69} & 
\cellcolor{mygray}{\textbf{30.23}} &
\cellcolor{mygray}{\textbf{17.49}} & \cellcolor{mygray}{\textbf{0.73}} & 
\cellcolor{mygray}{\textbf{20.78}}&
\cellcolor{mygray}{\textbf{18.60}} & \cellcolor{mygray}{\textbf{0.71}}\\
\hline
%\vspace{-1.1cm}
\end{tabular}

\end{center}
\end{table*}

\begin{figure*}[t]
\centering
\includegraphics[width=1\linewidth]{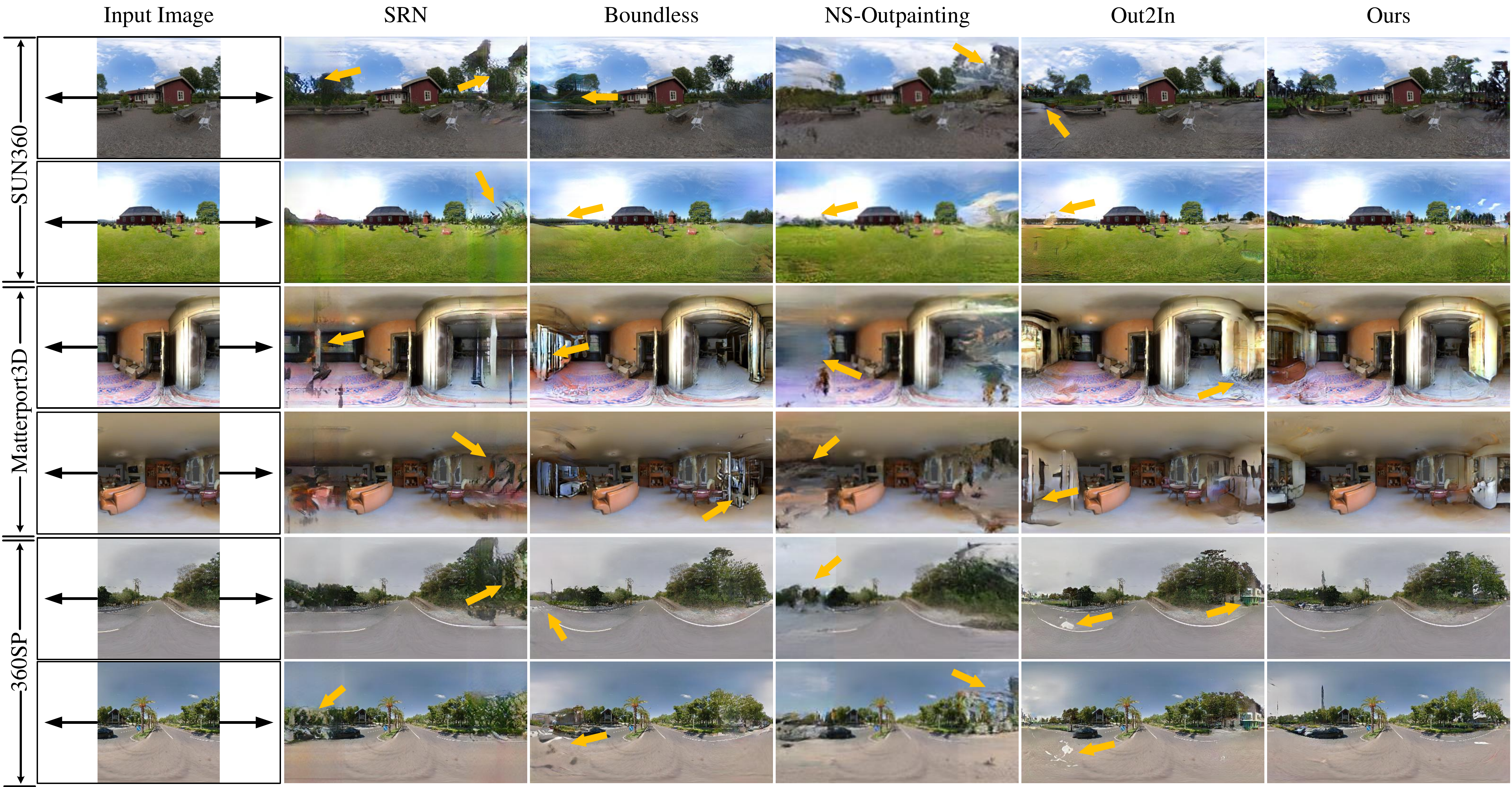}
\caption{Visual comparison of different methods on three datasets. We show the input image, the outpainting results of SRN \cite{13}, Boundless \cite{12}, NS-Outpainting \cite{14},  Out2In \cite{kim2021painting}, and Ours. The arrows highlight the artifacts and inferior generated regions.}
\label{fig:results}
%\vspace{-0.3cm}
\end{figure*}

Recall that in Transformers, 1D SPE can be described by the combination of sine and cosine functions of different frequencies. For the 360{\textdegree} panoramic image, a 2D SPE should be the concatenation of the 1D SPEs in azimuthal and polar angles:
\begin{equation}%\label{eq4}
[\underbrace{sin(\omega_1 \theta), cos(\omega_1 \theta), \cdots}_{azimuthal \ \ dimension}, \underbrace{sin(\omega_1 \varphi), cos(\omega_1 \varphi), \cdots}_{polar \ \ dimension} ],
\end{equation}
where $\omega_i = 1 / 10000^{2k/d}$ and $d$ indicates the half of the total encoding dimensions. 

In our Cylin-Painting, the cylinder-style convolution discards the positional information in azimuthal (horizontal) dimension as shown in Fig.~\ref{fig:SPE}. There are two problems that should be addressed before designing a new positional embedding: where to add and what to add. The experimental results show that the network's performance improves as the position information is gradually added to each convolutional layer. Note that we observe no performance gains when adding the position information into the decoder part. One possible reason is that the positional encoding has been transited from the encoder to the decoder via the skip connection. 

On the other hand, although SPE provides multi-frequency positional information, existing methods use it by only separately adding every single frequency to feature maps, which cannot effectively exploit the abundant multi-frequency information. 
A more reasonable approach is to adaptively aggregate different frequencies of SPE before the combination with image features.
Therefore, we design our learnable positional embedding as the following
\begin{equation}\label{eq4}
\mathcal{F}_{out}^{t}(\theta, \varphi) = \mathcal{K}^{t} * [\mathcal{F}_{in}(\theta, \varphi) \oplus \sum_{c=1}^{C^{\prime}}\mathcal{K}_{pe}^{(c)}*\mathcal{S}(\theta, \varphi, c)].
\end{equation}
This equation describes how to embed the learnable positional information into a cylinder convolutional layer based on Eq. \ref{eq3}. $\oplus$ means the element-wise addition. Moreover, a 1$\times$1 kernel $\mathcal{K}_{pe}^{(c)}$ learns to select and combine the effective position information from the hand-crafted SPE $\mathcal{S}$. As illustrated in Fig.~\ref{fig:SPE} (bottom), our learnable positional embedding can obtain an explicit, compact, and evenly-distributed 2D position map, which 
embodies both high-frequency/relative PE 
that describes the local spatial relationship of adjacent areas and low-frequency/absolute PE that offers a global positional encoding with a unique long-range relationship. More details about the proposed $\mathcal{S}$ will be discussed in Section \ref{s6.3}.

\section{Experiments}
\label{sec6}

\subsection{Experimental Settings}
\label{s6.1}
To comprehensively evaluate the performance of the proposed Cylin-Painting, we conduct extensive experiments on three 360{\textdegree} panoramic image datasets including outdoor and indoor scenes: SUN360 \cite{xiao2012recognizing}, Matterport3D \cite{chang2017matterport3d}, and 360SP \cite{chang2018generating}. For each dataset, we consider the image outpainting case with the resolution of $256\times256 \rightarrow 512\times256$.

We train our Cylin-Painting framework with settings similar to Boundless \cite{12}. In detail, we add the instance normalization \cite{38} after each convolutional layer. During training, we keep the generative adversarial scheme, and a discriminator network is applied to promote the realisticness of the results. The batch size of training is set to be $2$. The input image and output results are linearly scaled to the range $[-1, 1]$. Adam optimizer with learning rates $10^{-4}$ and $10^{-3}$ are adopted for the generator network and discriminator network, respectively. We use a loss function that combines a pixel-wise $\mathcal{L}_1$ loss and an adversarial loss $\mathcal{L}_{adv}$ to train the Cylin-Painting, in which the balance factors are empirically set to $\lambda_{gen} = 1$ and $\lambda_{adv} = 10^{-2}$. To combine the strengths of image inpainting and outpainting, the generator network leverages cylinder-convolutions but unintentionally omits the positional information. To this end, the learnable positional embedding is integrated into the generator network and assists it in creating more realistic and orderly structured content. For the discriminator, it is mainly responsible for the judgment of the image fidelity between the ground truth and the generated image, and thus we keep it in the same configuration as the commonly used one. The computation complexity of the proposed Cylin-Painting is listed as follows: running time per image: 0.0117s, MACs: 54.323G, and the numbers of parameters: 3.457M. All the evaluations are conducted on an NVIDIA GeForce RTX 2080 Ti GPU.

\subsection{Comparison Results}
\label{s6.2}
\noindent \textbf{Quantitative Evaluation:}
Following previous image outpainting methods, we use three metrics to evaluate the performance: the peak signal-to-noise ratio (PSNR), the structural similarity index (SSIM), and Fr\'echet Inception Distance (FID). Previous works \cite{12,13,bowen2021oconet} verified that PSNR and SSIM are not optimal metrics for evaluating conditional image generation tasks. Instead, FID is generally regarded as a crucial metric, which is marked with an underline in Table \ref{table:1}. We compare our approach with the state-of-the-art image outpainting methods: SRN \cite{13}, Boundless \cite{12}, NS-Outpainting \cite{14}, SpiralNet \cite{29}, Out2In \cite{kim2021painting}. Since the source codes of Out2In \cite{kim2021painting} are unavailable, we reproduce this work and achieve performance similar to that reported in the original paper. Then we compute three evaluation metrics using the difference between each extrapolated image and the ground truth image. As listed in Table \ref{table:1}, although our approach only slightly outperforms other methods in PSNR and SSIM, we lead all SOTA methods with a clear margin in terms of the FID metric. 

\noindent \textbf{Qualitative Evaluation:}
In Fig.~\ref{fig:results}, we show the results of the baseline methods and ours on three 360{\textdegree} panoramic image datasets. Intuitively, the results generated by SRN \cite{13} and NS-Outpainting \cite{14} suffer from poor extrapolated contents with unrelated objects and artifacts. Although more visually pleasing results are produced by Boundless \cite{12} and Out2In \cite{kim2021painting}, noticeable fractures of semantics occur near the boundary, as the spatial distance to the source region increases. Since the state-of-the-art Out2In \cite{kim2021painting} converts the challenging image outpainting problem into an inpainting one, the convolutional kernel is able to exact the features across original image boundaries. Thus, it achieves promising performance on image consistency beyond other baselines. Nevertheless, our method efficiently fuses different spatial arrangements of both outpainting and inpainting, showing more realistic image quality than Out2In \cite{kim2021painting}. 

\begin{figure}[htbp]
\centering
\includegraphics[width=1\linewidth]{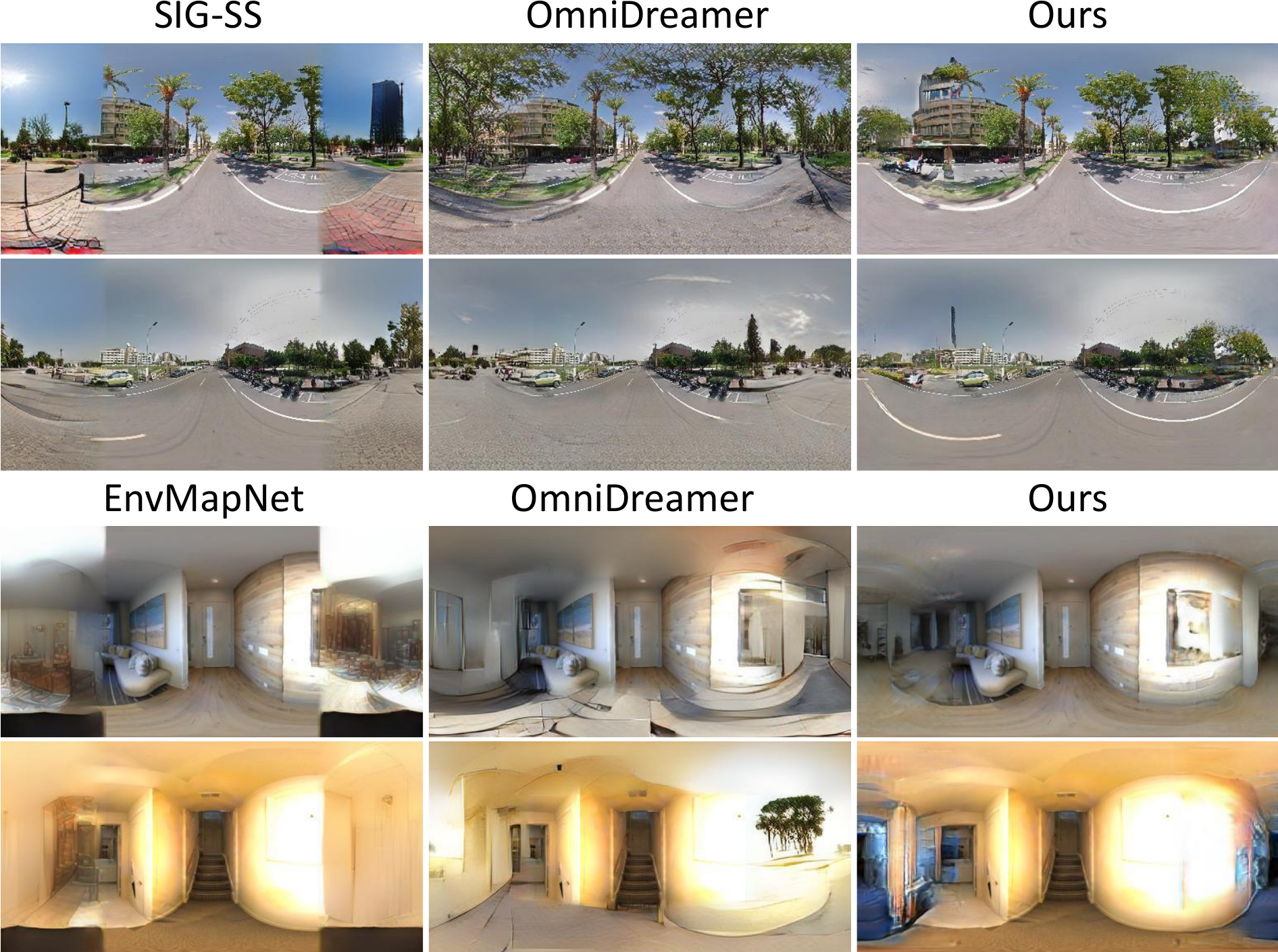}
\caption{Visual comparison on the state-of-the-art panoramic image outpainting works: SIG-SS~\cite{hara2021spherical},  EnvMapNet~\cite{somanath2021hdr}, and OmniDreamer~\cite{akimoto2022diverse}.}
\label{fig:pano-cp}
\end{figure}

\noindent \textbf{Panoramic Outpainting Evaluation:}
Beyond the comparison of the classical image completion methods, we further compared the proposed method with the state-of-the-art panoramic outpainting works. In particular, we have compared OmniDreamer~\cite{akimoto2022diverse}, SIG-SS~\cite{hara2021spherical}, and EnvMapNet~\cite{somanath2021hdr} with the proposed method, and the comparison results are shown in Figure~\ref{fig:pano-cp}. Following the comparison setting in OmniDreamer~\cite{akimoto2022diverse}, we also divided the scene into outdoor and indoor. Figure~\ref{fig:pano-cp} shows the state-of-the-art OmniDreamer~\cite{akimoto2022diverse} outperforms previous methods~\cite{hara2021spherical, somanath2021hdr} regarding the panoramic consistency and diversity of details. Instead, our method excels at content fidelity such as the complete lane lines and smooth floor, which is also verified by the quantitative evaluation in Table~\ref{tab:pano-cp}.

\begin{table}[htbp]\small
  \caption{Quantitative evaluation on the state-of-the-art panoramic image outpainting works: SIG-SS~\cite{hara2021spherical},  EnvMapNet~\cite{somanath2021hdr}, and OmniDreamer~\cite{akimoto2022diverse}, in terms of FID metric $\downarrow$.}
  \label{tab:pano-cp}
	 \centering
   \begin{tabular}{c|c|c|c|c}
    \hline
      & EnvMapNet & SIG-SS & OmniDreamer &
      Ours\\
    \hline
    \hline
    Indoor &43.05 &  - & 33.23 &
    30.23\\
    \hline
    Outdoor & - &  35.04 & 27.29&
    20.78\\
    
  \hline
\end{tabular}
\end{table}

\begin{figure}[t]
\centering
\includegraphics[width=1\linewidth]{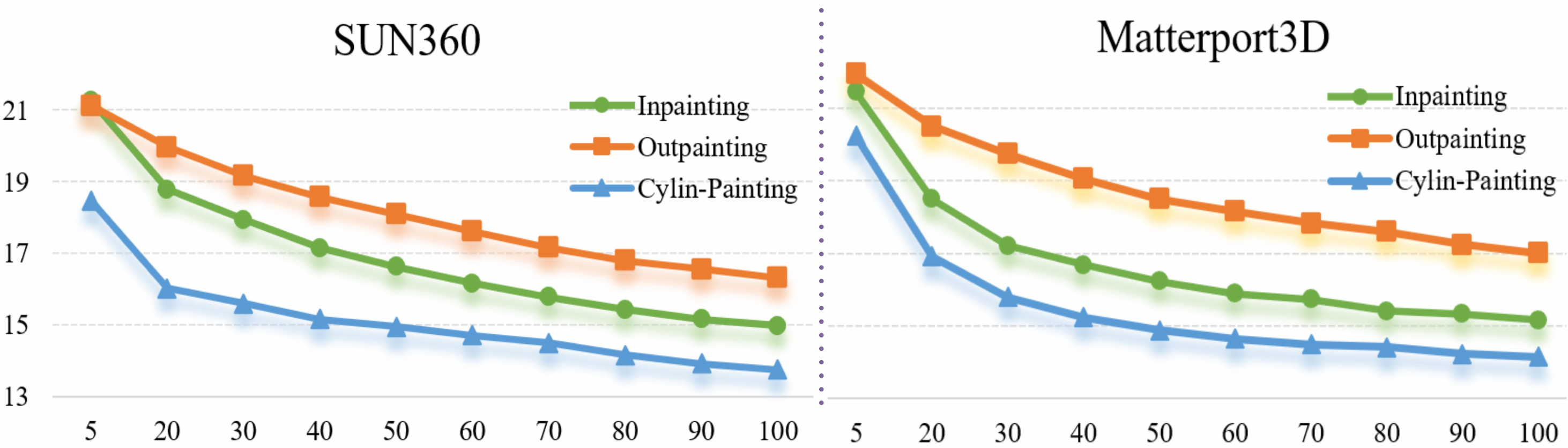}
\caption{Training loss comparisons of the inpainting, outpainting, and our Cylin-Painting on SUN360 \cite{xiao2012recognizing} and Matterport3D \cite{chang2017matterport3d}.}
\label{fig:loss}
%\vspace{-0.3cm}
\end{figure}

\begin{figure}[t]
\centering
\includegraphics[width=.9\linewidth]{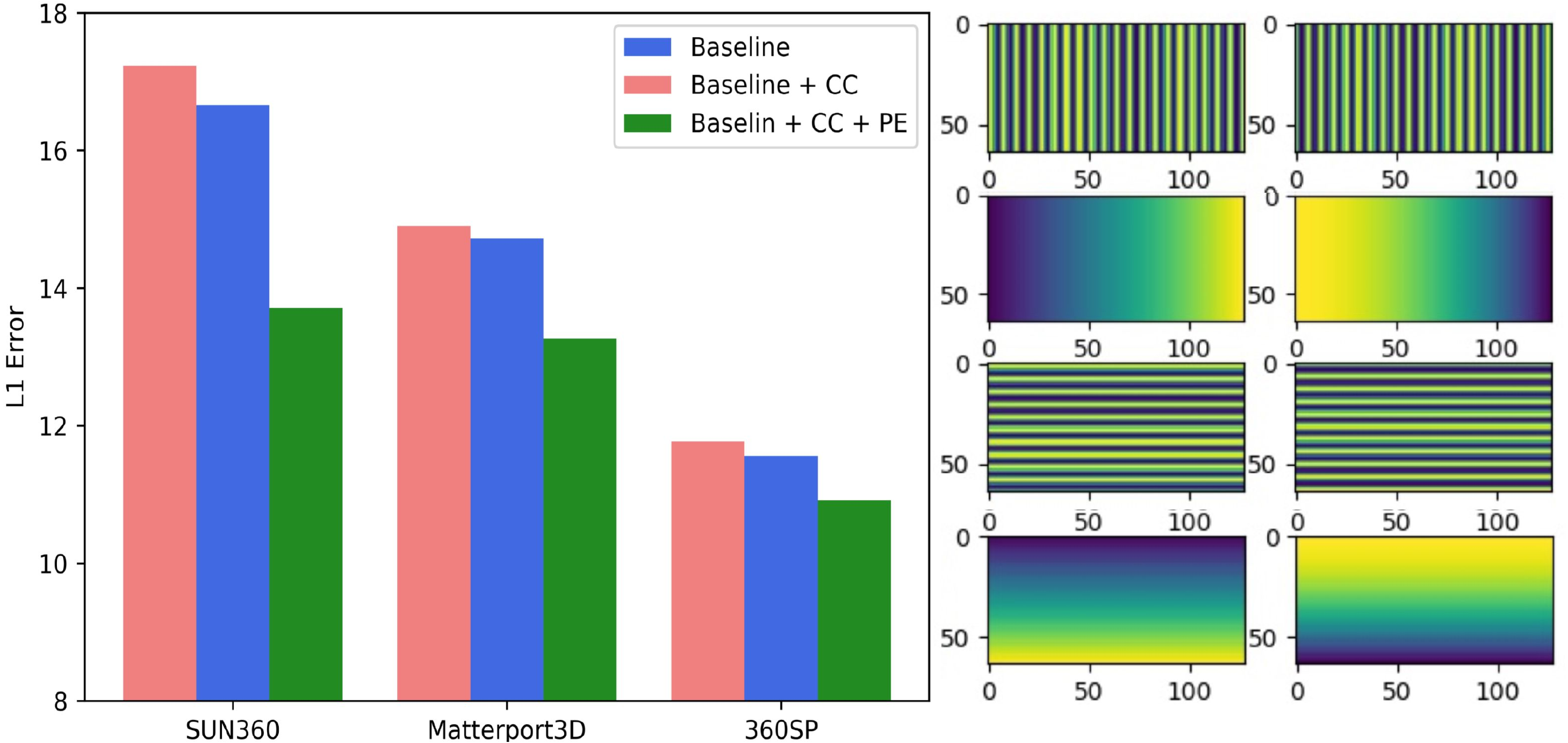}
\caption{Ablation study on positional encodings. Left: $\mathcal{L}_1$ errors ($\times 10^{-2}$) evaluated the generated results on three datasets, where CC = Cylinder Convolution, PE = Learnable Positional Embedding. Right: Samples of 2D positional encoding maps.}
\label{fig:bar}
%\vspace{-0.2cm}
\end{figure}

\begin{figure}[t]
\centering
\includegraphics[width=1\linewidth]{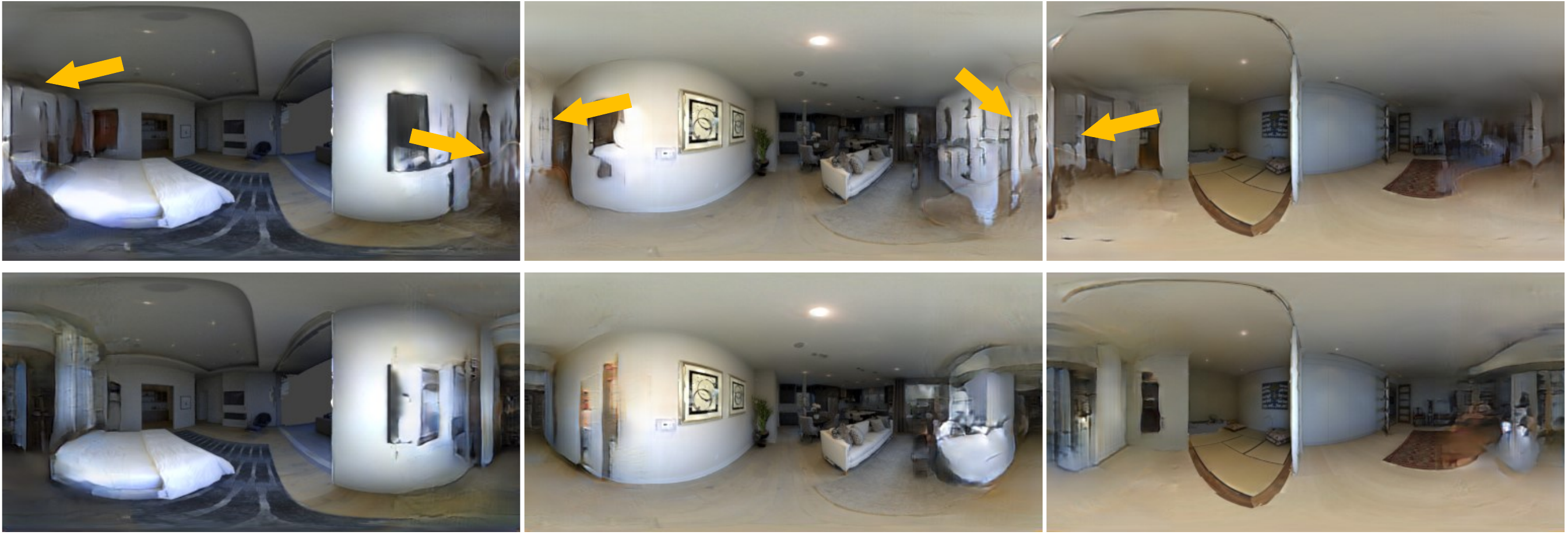}
\caption{Visual results of ablation study. Top: Ours w/o learnable positional embedding. Bottom: Ours w/ learnable positional embedding. The arrows highlight the disorder structures.}
\label{fig:ablation}
%\vspace{-0.3cm}
\end{figure}

\begin{table}\small
  \caption{Ablation study on the type of positional encodings,  $\mathcal{L}_1$ errors ($\times 10^{-2}$) evaluated of the generated results on three datasets, where RA = Relative Azimuthal, RP = Relative Polar, AA = Absolute Azimuthal, AP = Absolute Polar, and ALL denotes all channels of SPE. AP* indicates the AP positional encoding applied to vanilla convolutions.}
  \label{tab:SPE}
	 \centering
   \begin{tabular}{p{1.75cm}<{\centering}|p{0.6cm}<{\centering}|p{0.56cm}<{\centering}|p{0.6cm}<{\centering}|p{0.63cm}<{\centering}|p{0.6cm}<{\centering}||p{0.6cm}<{\centering}}
    \hline
    Types & RA & RP & AA & \textbf{AP} & ALL & AP* \\
    \hline
    \hline
    SUN360  & 18.69 & 16.51& 16.95 & \textbf{16.21} & 16.75  & 18.01 \\
    \hline
    Matterport3D & 15.95 & 15.01 & 14.53 & \textbf{13.85} & 14.36 & 16.41 \\
    \hline
    360SP & 11.60 & 11.78 & 11.68 & \textbf{11.46} & 11.70 & 11.58 \\

  \hline
\end{tabular}
%\vspace{-0.3cm}
\end{table}

\subsection{Ablation Study}
\label{s6.3}
To validate the effectiveness of different components in our approach, we conduct an ablation study in regard to spatial arrangement and positional encoding.

\noindent \textbf{Spatial Arrangement:}
As described in Section \ref{s3.2}, the difference between inpainting and outpainting essentially depends on how the source pixels contribute to the unknown locations under different spatial arrangements. As illustrated in Fig.~\ref{fig:loss}, we show the training loss curves of the inpainting, outpainting, and our Cylin-Painting. We can observe that by converting the outpainting to the inpainting, the introduced bidirectional flow is capable of relieving the learning burdens of the neural network, achieving better convergence on both two panoramic datasets. However, individually exploiting the inpainting or outpainting cannot trade-off their relative benefits, and thus limits the potential collaborations and further improvements. Our Cylin-Painting, by contrast, efficiently fuses the different spatial arrangements in inpainting and outpainting, promoting the learning model to converge faster and better.

\noindent \textbf{Positional Encoding:}
Although the cylinder convolution considers the prior in 360{\textdegree} panoramic images, it discards the azimuthal/horizontal positional information as shown in Fig.~\ref{fig:SPE}. Thus, it is hard for the network to build an accurate spatial relationship between the original contents and generated contents, leading to inferior outpainting results with disordered structures. In Fig.~\ref{fig:bar} (left), the baseline network equipped with the cylinder convolution (CC) even cannot beat the baseline. Further combined with our learnable positional embedding (PE), the network gains the crucial location-aware ability. Thus the final performance achieves impressive improvements beyond the baseline as depicted in Fig.~\ref{fig:ablation}.

On the other hand, there are various positional information provided by sinusoidal positional encoding (SPE). It is unreasonable to apply all information into limited convolutional layers. In Fig.~\ref{fig:bar} (right), we show eight channels from 128-dim SPE; they represent the 2D positional information from relative position (high-frequency) to absolute position (low-frequency), from polar dimension to azimuthal dimension. To determine which type of positional encoding is needed in CNNs, we conduct extensive experiments as listed in Table \ref{tab:SPE}. From these results, we can conclude that the absolute polar (AP) positional encoding allows the best benefit for the learning model. This experimental result supports the analysis in Section \ref{s5.4}. We also found that when using all channels of SPE (ALL) or applying the AP positional encoding into vanilla convolutions (AP*), the performance degenerates inferior to the baseline. Such a phenomenon indicates that CNNs are sensitive to hybrid and ambiguous positional information.

 \begin{figure}[!t]
    \centering
    \includegraphics[width=\linewidth]{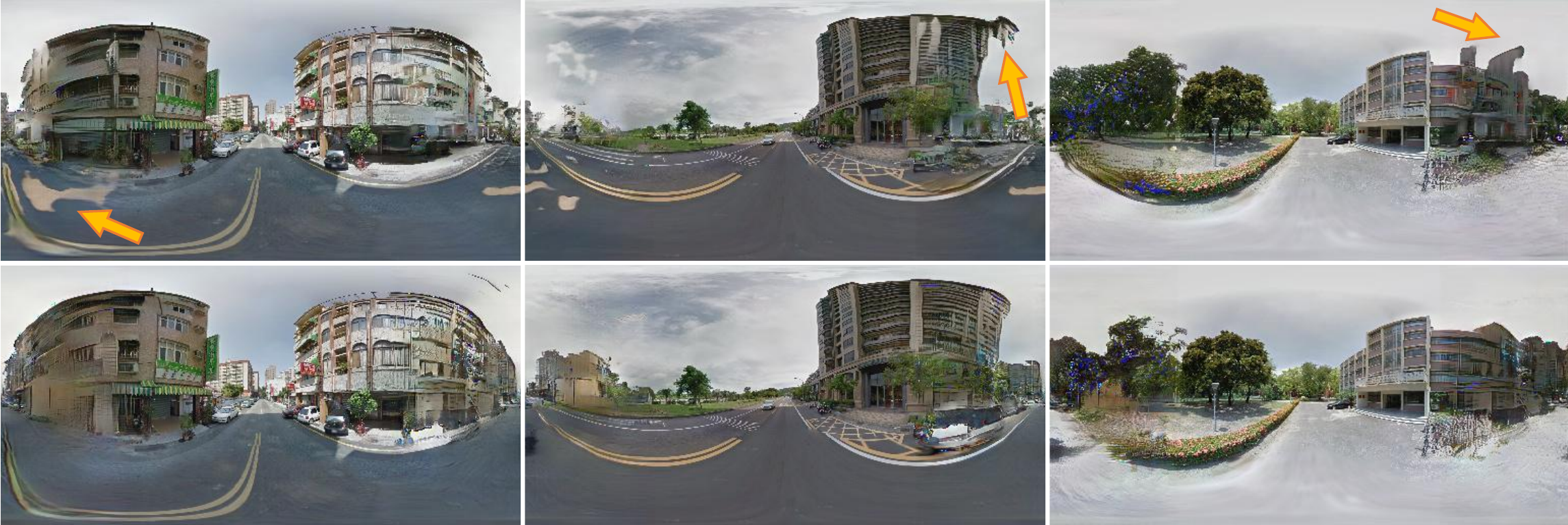}
    \caption{Comparison of the panoramic-oriented positional encoding (top) and the proposed positional embedding (bottom) for panoramic image outpainting.}
    \label{fig:PE_cp}
\end{figure}

We also incorporate the panoramic-oriented positional encodings~\cite{jiang2022lgt, li2023panoramic, zheng2023both} into our framework. In Figure~\ref{fig:PE_cp}, while the panoramic-oriented positional encoding enforces the panoramic characteristics in the outpainting results, it would introduce some unreasonable structures on objects, such as the ground and building. Compared to directly using the positional encoding strategy in Transformer-based solutions, our learnable positional embedding is customized for CNNs, which learns to select gainful positional information in terms of different scenarios.    

\begin{table}\small
  \caption{Quantitative evaluation of depth estimation for panoramic images. \textbf{\textcolor{red}{Red}} text indicates the best and \uline{\textcolor{blue}{blue}} text indicates the second-best performing method.}
  \label{tab:Depth}
	 \centering
   \begin{tabular}{p{1.17cm}<{\centering}|p{0.7cm}<{\centering}|p{0.75cm}<{\centering}|p{0.7cm}<{\centering}||p{0.65cm}<{\centering}|p{0.75cm}<{\centering}|p{0.7cm}<{\centering}}
    \hline
    Methods  Metrics & FRCN \cite{laina2016deeper} & FRCN + CC & FRCN + CC + PE & HR \cite{wang2020deep} & HR + CC & HR + CC + PE \\
    \hline
    \hline
    RMSE $\downarrow$ & \uline{\textcolor{blue}{0.6549}} & 0.6890 & \textbf{\textcolor{red}{0.6476}} & \uline{\textcolor{blue}{0.36}} & 0.37  & \textbf{\textcolor{red}{0.34}} \\
    \hline
    $\delta_{1}\uparrow$ & \uline{\textcolor{blue}{0.7359}} & 0.7353 & \textbf{\textcolor{red}{0.7674}} & \uline{\textcolor{blue}{0.90}} & 0.90 & \textbf{\textcolor{red}{0.92}} \\
    \hline
    $\delta_{2}\uparrow$ & 0.9075 & \uline{\textcolor{blue}{0.9125}} & \textbf{\textcolor{red}{0.9213}} & 0.97 & \uline{\textcolor{blue}{0.98}} & \textbf{\textcolor{red}{0.99}} \\
    \hline
    $\delta_{3}\uparrow$ & 0.9634 & \uline{\textcolor{blue}{0.9660}} & \textbf{\textcolor{red}{0.9715}} & \uline{\textcolor{blue}{0.99}} & 0.99 & \textbf{\textcolor{red}{0.99}} \\

  \hline
\end{tabular}
%\vspace{-0.2cm}
\end{table}

\begin{figure}[t]
\centering
\includegraphics[width=1\linewidth]{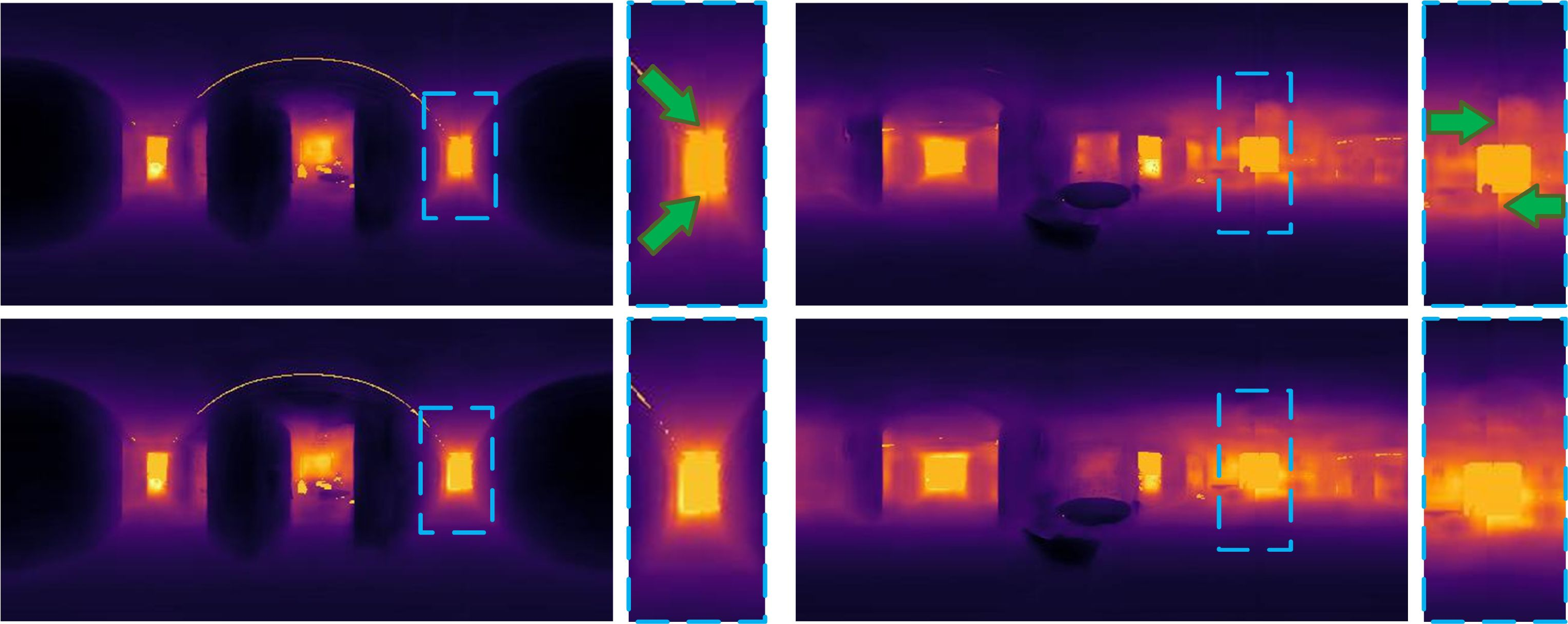}
\caption{Visual depth estimation comparisons of HR-Net \cite{wang2020deep} (top) and the improved baseline benefited by our solution (bottom). The arrows highlight the seamed and inconsistent content.}
\label{fig:depth}
%\vspace{-0.2cm}
\end{figure}

\begin{table}\small
  \caption{Quantitative evaluation of object detection for panoramic images. We employ the classical Faster-RCNN as the baseline model. \textbf{\textcolor{red}{Red}} text indicates the best and \uline{\textcolor{blue}{blue}} text indicates the second-best performing method.}
  \label{tab:OD}
	 \centering
   \begin{tabular}{c|c|c|c}
    \hline
      & Baseline & Baseline+CC & Baseline+CC+PE \\
    \hline
    \hline
    mAP $\uparrow$ &34.98 &  \uline{\textcolor{blue}{35.01}} & \textbf{\textcolor{red}{35.45}}\\

  \hline
\end{tabular}
%\vspace{-0.3cm}
\end{table}

\begin{figure}[t]
\centering
\includegraphics[width=1\linewidth]{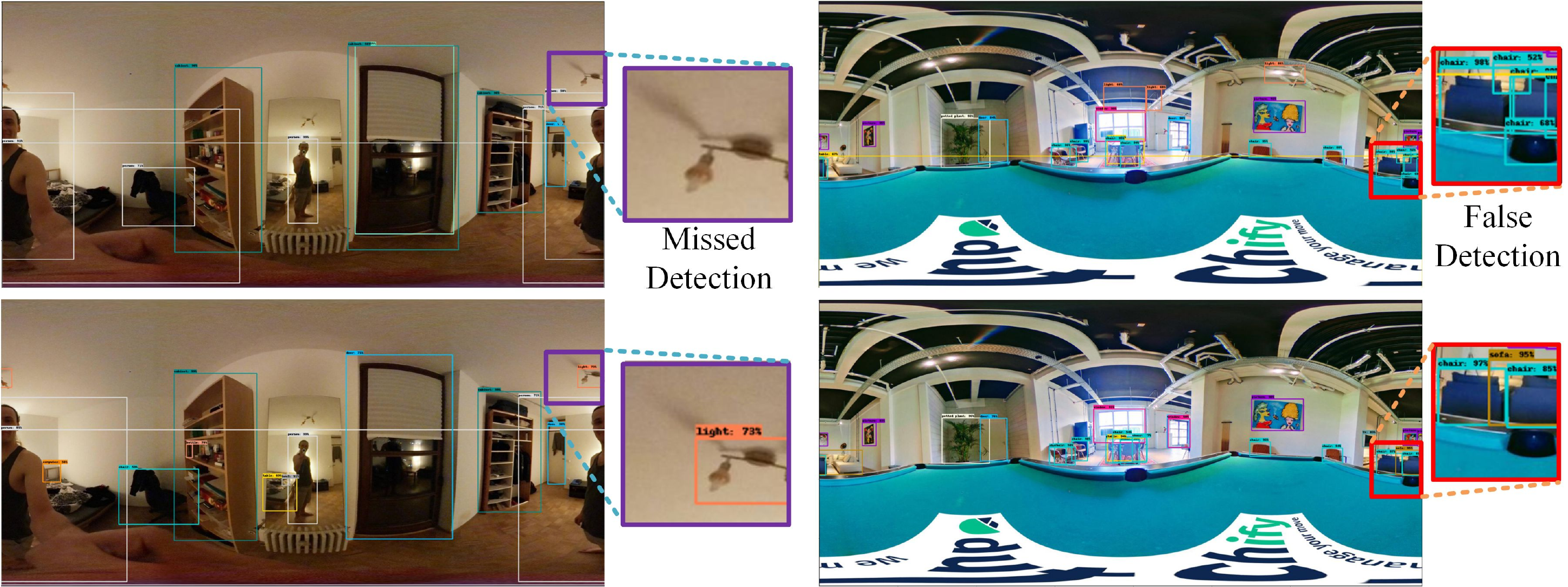}
\caption{Object detection comparisons of Faster-RCNN \cite{ren2015faster} (top) and the improved baseline by our solution (bottom). The purple rectangle and red rectangle highlight the comparisons of missed object detection and false object detection, respectively.}
\label{fig:detection}
%\vspace{-0.4cm}
\end{figure}

\subsection{Applications to More Panoramic Vision Tasks}
\label{s6.4}
While designed for panoramic image outpainting, our method can be effectively extended to other panoramic vision tasks, including high-level and low-level tasks. In the implementation, we leverage the proposed cylinder-style convolution (CC) and learnable positional embedding (PE) as a plug-and-play module for prevalent learning models.

\noindent \textbf{Panoramic Depth Estimation:}
We choose a classical FRCN \cite{laina2016deeper} and a state-of-the-art HR-Net \cite{wang2020deep} to conduct this experiment on the Stanford 2d3d dataset. Standard evaluation metrics are used for the quantitative evaluation--- root mean square error of linear metric (RMSE), and three relative accuracy metrics $\delta_{1}$, $\delta_{2}$ and $\delta_{3}$, defined for an accuracy $\delta_{n}$, as the fraction of pixels where the relative error is within a threshold of $1.25 ^ n$. Table \ref{tab:Depth} illustrates the quantitative results of the baseline networks, in comparison with the improved version with our cylinder-style convolution (CC) and learnable positional embedding (PE). As mentioned in Section \ref{sec4}, while the cylinder convolution enables a seamless perception for panoramic images, it discards the polar positional encoding on the feature maps. As a result, directly exploiting the cylinder convolution into a panoramic depth estimation network decreases the accuracy of depth maps. When collaborating with our learned positions, the final version (baseline + CC + PE) significantly outperforms the baseline network on all metrics especially in the classical FRCN \cite{laina2016deeper}. For the qualitative evaluation, we rotate the estimated depth map with 90{\textdegree} as shown in Fig.~\ref{fig:depth}. We can observe that our solution can make the depth values more consistent across two original boundaries.

\noindent \textbf{Panoramic Object Detection:} The classical Faster-RCNN \cite{ren2015faster} is selected as the baseline network and the detection results are illustrated in Fig.~\ref{fig:detection}. While excellent performance is achieved in the central region, the prevalent 2D detector fails to learn the key prior of the 360{\textdegree} panoramic image, i.e., the consistent spatial arrangement, ignoring the correspondence between the left and right boundaries. Consequently, the objects located at the boundary are difficult to detect accurately, as shown in Fig.~\ref{fig:detection} (top). By contrast, our cylinder-style convolution learns the semantic features on a consecutive circular surface, and thus helps to detect the missed objects and eliminate the false objects in Fig.~\ref{fig:detection} (bottom). Besides, our learnable positional embedding provides a clear and compact spatial location for the network, which further improves the detection performance with $+ 0.47$ mAP improvement beyond the baseline as listed in Table \ref{tab:OD}. Although the manually cyclical shift to the panoramic image helps detect the object at boundaries, it can lead to the discontinuity of other objects. Instead, our method perceives the whole scene on a seamless cylinder, thereby all objects can be completely detected without any rotations.

\begin{figure}[htbp]
\centering
\includegraphics[width=1\linewidth]{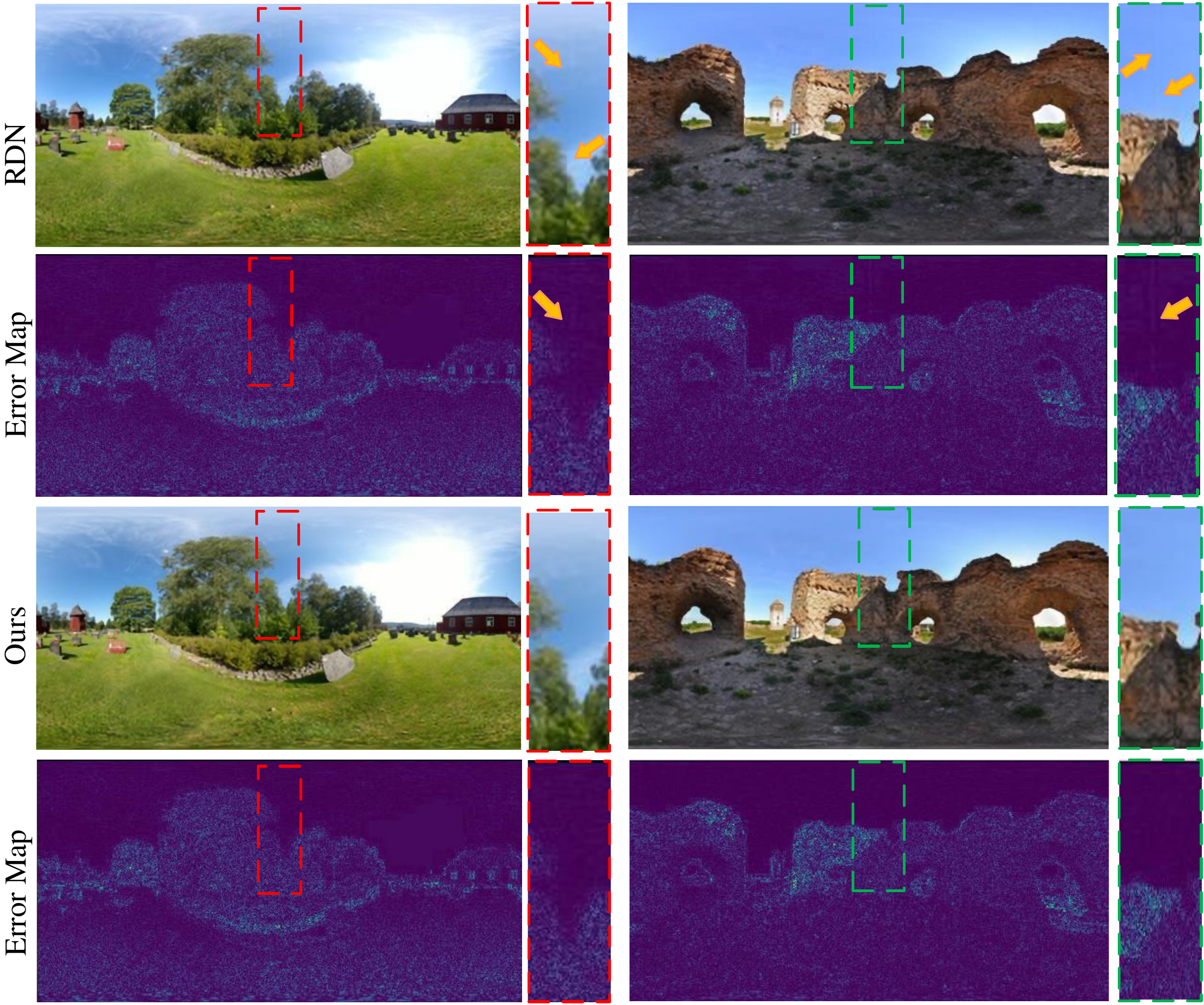}
\caption{Visual image super-resolution comparisons of RDN \cite{zhang2018residual} and the improved baseline by our solution. The yellow arrows highlight the seamed and inconsistent content. We also show the error map between super-resolution results and ground truth, in which the arrows point to the seams.}
\label{fig:SR}
%\vspace{-0.5cm}
\end{figure}

\noindent \textbf{Panoramic Image Super-Resolution:} We choose the RDN \cite{zhang2018residual} and EDSR \cite{lim2017enhanced} as two baseline networks. As listed in Table \ref{tab:SR}, by replacing the vanilla convolution with cylinder-style convolution, the networks' performance on content enhancement gains intuitive improvements without extra parameters introduced. Moreover, our solution enables a seamless panoramic image super-resolution as shown in Fig.~\ref{fig:SR}, in which the stitched part of the left and right boundaries display smooth and consistent appearances. We also show the error map between super-resolution results and ground truth, in which the arrows point to the seams that exist in baseline methods. However, Table \ref{tab:SR} demonstrates that further combinations with the positional embedding (PE) decrease the network's performance in terms of PSNR and SSIM.

\begin{table}[htbp]\small
  \caption{Quantitative evaluation of panoramic image super-resolution ($\times 2$) networks. \textbf{\textcolor{red}{Red}} text indicates the best and \uline{\textcolor{blue}{blue}} text indicates the second-best performing method.}
  \label{tab:SR}
	 \centering
   \begin{tabular}{p{1.18cm}<{\centering}|p{0.7cm}<{\centering}|p{0.75cm}<{\centering}|p{0.7cm}<{\centering}||p{0.65cm}<{\centering}|p{0.75cm}<{\centering}|p{0.7cm}<{\centering}}
    \hline
    Methods  Metrics & RDN \cite{zhang2018residual} & RDN + CC & RDN + CC + PE & EDSR \cite{lim2017enhanced} & EDSR + CC & EDSR + CC + PE \\
    \hline
    \hline
    PSNR $\uparrow$ & \uline{\textcolor{blue}{32.59}} & \textbf{\textcolor{red}{32.68}} & 32.58 & \uline{\textcolor{blue}{32.25}} & \textbf{\textcolor{red}{32.43}}  & 30.62 \\
    \hline
    SSIM $\uparrow$ & \uline{\textcolor{blue}{0.917}} & \textbf{\textcolor{red}{0.922}} & 0.917 & \uline{\textcolor{blue}{0.902}} & \textbf{\textcolor{red}{0.904}} & 0.890 \\

  \hline
\end{tabular}
%\vspace{-0.3cm}
\end{table}

\noindent \textbf{Panoramic Semantic Segmentation:}
We select the state-of-the-art CBFC~\cite{zheng2023complementary} as the baseline since its main framework is also CNNs same as our studied case. We integrate the proposed positional embedding strategy into the baseline~\cite{zheng2023complementary}. The resulting model achieves an mIoU of 53.9, which is only slightly better than the original model's 53.8.
We hypothesize that this may be attributed to the crucial role of recovering detailed structures in segmentation, akin to the case of super-resolution, which diminishes the location advantages provided by positional encoding. This also indicates that further analysis and discussions are needed to fully address the complex relationship between these factors in segmentation tasks, which will be an interesting direction for future research.

In summary, different combinations of our solutions can effectively help various panoramic vision tasks. We also conclude that the combination of our solution depends on the type of vision task. For high-level vision tasks such as depth estimation and object detection, our cylinder-style convolution with the learnable positional embedding shows the best performance. The main reason is the high-level task requires clear and accurate position information, which facilitates building the spatial relationship of different semantics. By contrast, low-level vision tasks such as image super-resolution pay more attention to the enhancement/reconstruction of local details. In addition, the positional embedding and zero padding restrain the weak response on the feature map, violating the principle of high content fidelity. Therefore, using only the cylinder-style convolution can boost the panoramic super-resolution. We hope our solution and empirical study could promote the development of subsequent 360{\textdegree} panoramic vision tasks.

\subsection{Limitation Discussion and Future Work}
Our method achieves promising performance on the panoramic image outpainting and shows generalizability to other panoramic vision tasks. However, the idea of incorporating the inpainting and outpainting is hard to be applied to normal 2D images, since the inconsistent contents appear in their boundaries. In future work, we plan to address this limitation by devising a truncated information flow, to enable a general solution for image completion. Moreover, it is interesting to apply our method to handle multiple inputs. We believe our method can be flexibly extended to such a case by masking the original panorama into some individual regions. Then the proposed Cylin-Painting is finetuned to learn this new mapping relationship.

\section{Conclusion}\label{sec7}
In this work, we target the 360{\textdegree} panoramic image outpainting task. It is more challenging than image inpainting due to the one-side constraint. Previous methods convert the image outpainting problem to an inpainting one without extensively analyzing the difference and relationship between these two technologies. We close this gap by showing that their difference depends on how the source pixels contribute to the unknown locations under different spatial arrangements. To combine their benefits, we present a Cylin-Painting framework to efficiently fuse the different spatial arrangements on a cylinder. Moreover, we further equip the cylinder-style convolution with a learnable positional embedding, which essentially improves the extrapolation results. While targeting the image outpainting, our solution can serve as a plug-and-play module and flexibly extend to other panoramic vision tasks. Furthermore, we provide a comprehensive analysis of the zero padding and positional encoding in CNNs. This study shows possible effort directions for subsequent 360{\textdegree} panoramic vision tasks. 

\section*{ACKNOWLEDGEMENT}
We would like to thank Zhijie Shen for valuable discussions and thank Keyao Zhao, and Zishuo Zheng for their help in experiments.

%\normalem
\bibliographystyle{ieeetr}

\end{document}